\documentclass[]{fairmeta}

\usepackage[utf8]{inputenc}
\usepackage{subcaption}

\usepackage[T1]{fontenc}    %
\usepackage{url}            %
\usepackage{hyperref}       %
\usepackage{booktabs}       %
\usepackage{amsfonts}       %
\usepackage{amssymb}
\usepackage{nicefrac}       %
\usepackage{microtype}      %
\usepackage[dvipsnames]{xcolor}

\usepackage{nicematrix}
\usepackage{algorithm}
\usepackage{algpseudocode}
\usepackage{amsthm}
\usepackage[normalem]{ulem}
\usepackage{wrapfig}
\usepackage{graphicx}
\usepackage{cancel}
\usepackage{multirow}
\usepackage{multicol}
\usepackage{amssymb}
\usepackage{cleveref}
\usepackage{minted}  
\usepackage{listings}

\usepackage{siunitx}
\usepackage{cleveref}
\usepackage{chngcntr}
\usepackage{csquotes}
\usepackage{enumitem}

\lstdefinestyle{simple}{
  basicstyle=\ttfamily\small,     %
  columns=fullflexible,
  backgroundcolor=\color{gray!8}, %
  frame=single,                    %
  rulecolor=\color{black!30},      %
  numberstyle=\tiny\color{gray},   %
  keywordstyle=\color{blue},       %
  commentstyle=\color{green!40!black}, %
  stringstyle=\color{red!70!black},    %
  showstringspaces=false,          %
  tabsize=2,                       %
  breaklines=true,                 %
  breakatwhitespace=true,          %
  captionpos=b,                     %
  xleftmargin=3.4pt,
  xrightmargin=3.4pt
}

\definecolor{cwm_keyword_color}{HTML}{0064E0}
\definecolor{cwm_string_color}{HTML}{009B9B}
\definecolor{cwm_string_color_bright}{HTML}{C80A28}
\definecolor{cwm_comment_color}{HTML}{D75FAA}

\lstdefinestyle{simple_small}{
  basicstyle=\fontsize{6.5}{7}\ttfamily, %
  columns=fullflexible,
  keywordstyle=\color{cwm_keyword_color},
  commentstyle=\color{cwm_comment_color},
  stringstyle=\color{cwm_string_color},
  breaklines=true,
  frame=single,
  escapeinside={(*@}{@*)}
}

\lstdefinestyle{simple_small_small}{
  basicstyle=\fontsize{6}{6.5}\ttfamily, %
  columns=fullflexible,
  keywordstyle=\color{cwm_keyword_color},
  commentstyle=\color{cwm_comment_color},
  stringstyle=\color{cwm_string_color},
  breaklines=true,
  frame=single,
  escapeinside={(*@}{@*)}
}

\definecolor{cwm_think_color}{HTML}{AFD7FF}
\lstdefinestyle{cwm_think}{
  framerule=0pt,
  columns=fullflexible,
  basicstyle=\ttfamily\tiny,     %
  backgroundcolor=\color{cwm_think_color}, %
  showstringspaces=false,          %
  breaklines=true,                 %
  xleftmargin=3.4pt,
  xrightmargin=3.4pt,
  breakindent=0pt
}
\definecolor{cwm_prompt_color}{HTML}{A8E6CF}
\lstdefinestyle{cwm_prompt}{
  framerule=0pt,
  columns=fullflexible,
  basicstyle=\ttfamily\tiny,     %
  backgroundcolor=\color{cwm_prompt_color}, %
  showstringspaces=false,          %
  breaklines=true,                 %
  xleftmargin=3.4pt,
  xrightmargin=3.4pt,
  breakindent=0pt
}

\definecolor{cwm_act_color}{HTML}{D2D2FF}
\lstdefinestyle{cwm_act}{
  framerule=0pt,
  columns=fullflexible,
  basicstyle=\ttfamily\tiny,     %
  backgroundcolor=\color{cwm_act_color}, %
  showstringspaces=false,          %
  breaklines=true,                 %
  xleftmargin=3.4pt,
  xrightmargin=3.4pt,
  breakindent=0pt
}

\definecolor{cwm_obs_color}{HTML}{FFDCB9}
\lstdefinestyle{cwm_obs}{
  framerule=0pt,
  columns=fullflexible,
  keepspaces=true,
  basicstyle=\ttfamily\tiny,     %
  backgroundcolor=\color{cwm_obs_color}, %
  showstringspaces=false,          %
  breaklines=true,                 %
  xleftmargin=3.4pt,
  xrightmargin=3.4pt,
  breakindent=0pt
}

\usepackage{placeins}

\title{Towards a Neural Debugger for Python}

\author{Maximilian Beck$^{1,*}$}
\author{Jonas Gehring$^2$}
\author{Jannik Kossen$^2$}
\author{Gabriel Synnaeve$^2$}

\affiliation{$^1$Johannes Kepler University Linz, Institute for Machine Learning}
\affiliation{$^2$Meta FAIR CodeGen Team}
\contribution{$^*$Work done during an internship at Meta FAIR}

\date{March 2026}

\abstract{
Training large language models (LLMs) on Python execution traces grounds them in code execution and enables the line-by-line execution prediction of whole Python programs, effectively turning them into \emph{neural interpreters}~\citep{codgenteam2025_cwm}. 
However, developers rarely execute programs step by step; instead, they use debuggers to stop execution at certain breakpoints and step through relevant portions only while inspecting or modifying program variables.
Existing neural interpreter approaches lack such interactive control.
To address this limitation,
we introduce \emph{neural debuggers}: language models that emulate traditional debuggers, supporting operations such as stepping into, over, or out of functions, as well as setting breakpoints at specific source lines.
We show that neural debuggers---obtained via fine-tuning large LLMs or pre-training smaller models from scratch---can reliably model both forward execution (predicting future states and outputs) and inverse execution (inferring prior states or inputs) conditioned on debugger actions.
Evaluated on CruxEval, our models achieve strong performance on both output and input prediction tasks, demonstrating robust conditional execution modeling.
Our work takes first steps towards future agentic coding systems in which neural debuggers serve as a world model for simulated debugging environments, providing execution feedback or enabling agents to interact with real debugging tools. 
This capability lays the foundation for more powerful code generation, program understanding, and automated debugging.
}

\begin{document}
\maketitle

\section{Introduction}
Debugging is the process of isolating and correcting mistakes in computer programs~\citep{johnson1982debuggingglossary}.
It is a fundamental task in software engineering that is often considered separate from writing programs~\citep{whitington2024debuggingfunctionalprogramsinterpretation}. 
A debugger is a collection of software tools to aid debugging~\citep{johnson1982debuggingglossary} that allows developers to inspect and control program execution through actions such as stepping into or over function calls, setting breakpoints, or returning from functions. 
By observing how program states evolve line by line, developers can localize faults, understand control flow, and reason about program correctness. 

Recent advances in large language models (LLMs) trained on massive code corpora have shown remarkable capabilities in code generation, completion, and repair~\citep{chen2021evaluatinglargelanguagemodels, lozhkov2024starcoder2stackv2, hui2024qwen25codertechnicalreport, guo2024deepseekcoderlargelanguagemodel, codegemmateam2024codegemmaopencodemodels, rozière2024codellamaopenfoundation}. 
In the coding domain, LLMs have evolved from supportive programmer tools---such as code completion---to agents writing complete codebases autonomously and are increasingly used to assist developers in debugging or fixing software bugs~\citep{handa2025economictasksperformedai}. 
However, open-source and academic models primarily reason over \emph{static} code and are not explicitly grounded in program execution. 

Therefore, several approaches expose language models to program execution data. 
One line of work incorporates feedback such as test results~\citep{gehring2025rlef}, error messages~\citep{zheng2025makeslargelanguagemodels}, or runtime/shell outputs~\citep{yang2024sweagentagentcomputerinterfacesenable, wei2025swerladvancingllmreasoning} into reinforcement learning~\citep{shao2024deepseekmathpushinglimitsmathematical} or iterative generation loops~\citep{chen2023teachinglargelanguagemodels, ni2024nextteachinglargelanguage}. 
Another complementary direction uses \emph{execution traces}, i.e., records of variable states and control flow transitions, to directly teach models the semantics of code execution~\citep{nye2021_showyourwork,liu2023_code_exec,armengol2025_what,codgenteam2025_cwm} in order to improve code generation and reasoning tasks such as verification, testing, and repair.
Specifically, these approaches train language models on complete execution traces, which enable the line-by-line prediction of whole programs, effectively turning these models into \emph{neural interpreters}. 
However, this overlooks how developers actually use debuggers to interact with programs and resolve bugs. 
Instead of executing programs strictly sequentially, they pause execution at certain breakpoints and step only through relevant parts while inspecting or modifying program variables.
Previous works fall short of modeling this interactive, non-sequential debugging behavior.

To address this shortcoming, we introduce \emph{neural debuggers}: neural networks that can serve as simulated debugging environments for Python programs. 
Specifically, given the program code as context, neural debuggers are language models trained to predict the line-by-line execution of a computer program, conditioned on typical debugger actions such as \texttt{step\_into}, \texttt{step\_over}, \texttt{breakpoint}, or \texttt{step\_return}. 

Unlike conventional debuggers, neural debuggers can simulate program execution even for non-executable or partially specified programs, making them applicable to debugging, testing, or synthesis scenarios where access to full execution environments is unavailable or restricted~\citep{zhuo2025_cyberzerotrainingcybersecurityagents}.
Moreover, traditional debuggers require re-executing the program after code or state modifications, leading to slow iteration cycles.
In contrast, neural debuggers enable efficient reinitialization of both program state and execution context through prompting and regeneration.
Beyond standard forward execution, our neural debuggers also support approximate \emph{inverse execution}:
given an arbitrary program state, they can infer or sample plausible preceding program states or inputs. 
In agentic coding systems, neural debuggers can function as learned world models of debugging environments. 
They can simulate execution feedback and external interactions—such as file reads, API requests, or operating system calls---or allow the agent to interact with real debugging environments.
Incorporating neural debuggers in this manner can augment existing coding systems with debugging and planning abilities, facilitating more effective code generation, comprehension, and repair.

\begin{figure}[tbp!]
    \centering
    \includegraphics[width=\linewidth]{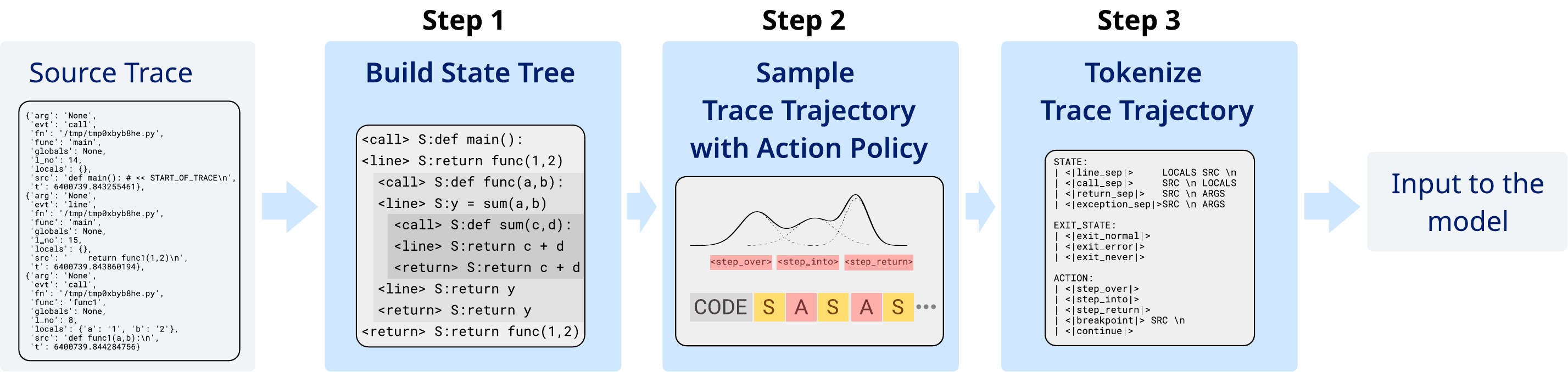}
    \caption{Neural Debugger Data Pipeline. 
    Our pipeline prepares training data for neural debuggers by transforming stack-frame sequences recorded via \texttt{sys.settrace} in three steps: (1) we construct a state tree (Section~\ref{sec:debugger_mdp}) from frame events; (2) we sample trajectories by traversing the state tree using a data-generating action policy; and (3) we tokenize each trajectory using our formal neural debugger language grammar (Section~\ref{sec:forward_inverse_trace_format}).
    }
    \label{fig:data_pipeline}
\end{figure}

In this paper, we take the first steps towards this vision by introducing neural debuggers as a Markov Decision Process (MDP), where each state represents the current program location and variable values, and transitions correspond to debugger actions that traverse a call-stack tree reconstructed from execution traces. 
We introduce a data pipeline (Figure~\ref{fig:data_pipeline}) that supports both forward and inverse execution prediction of Python programs.
The pipeline takes execution traces as input, samples debugger trajectories, and tokenizes them into a structured format compatible with standard off-the-shelf language models. 

We finetune \SI{32}{B}-parameter and pre-train \SI{1.8}{B}-parameter LLMs on this data to obtain neural debuggers that accurately predict program execution conditioned on debugger actions. 
Our \SI{32}{B} parameter neural debugger consistently achieves forward next state prediction accuracies beyond \SI{90}{\percent} across key actions: step into, step over, step return, and breakpoint.
In addition, neural debuggers exhibit strong transfer to CruxEval output and input prediction tasks~\citep{gu2024cruxeval}, demonstrating their enhanced code understanding and interpretation capabilities.
In particular, our \SI{1.8}{B}-parameter neural debugger LLM, trained from scratch on \SI{150}{B} tokens, attains CruxEval input and output pass@1 scores of \SI{53.6}{} and \SI{57.7}{}, respectively, while our finetuned \SI{32}{B}-parameter neural debugger LLM reaches scores of \SI{66.5}{} and \SI{83.2}{}, respectively.

\newpage
In summary, in this work we make the following contributions:
\begin{itemize}
    \item We introduce neural debuggers: language models that can predict forward and inverse execution of Python programs conditioned on source code and debugger actions.
    \item We describe a data pipeline for building neural debugger models by starting from pre-trained LLMs via finetuning or integration into pre- or mid-training data mixes. 
    \item We empirically show that neural debuggers achieve accurate intermediate state predictions and strong overall execution prediction performance.
\end{itemize}

\section{Related work}
\label{sec:related_work}
The problem of training neural networks to simulate computer program execution, like learned interpreters, is a long standing problem and has been studied with domain-specific architectures~\citep{zaremba2014learningexecute, reed2016neuralprogrammerinterpreters, wang2020learningsemanticprogramembeddings, bieber2020_learning_attention_graph} and more recently with Transformer-based LLMs. %
For example, \citet{nye2021_showyourwork} train Transformer models to predict intermediate states computed and source lines visited during the execution of Python functions.
They refer to this approach as \enquote{scratchpad tracing}, which they find to outperform the direct prediction of function outputs.
\citet{armengol2025_what} compare different scratchpad strategies for storing intermediate computations by training and evaluating models on different execution trace granularities, i.e., line and instruction level.
Their proposed dynamic scratchpads, in which the model updates a single self-contained scratchpad instance, produce more accurate predictions and improve performance on CruxEval output prediction tasks.
\citet{bieber2022staticpredictionruntimeerrors} introduce a dataset of Python runtime errors and train a Graph Neural Network on this dataset to predict statically whether a program will encounter a runtime error when it is executed. %
\citet{liu2023_code_exec} study code execution capabilities with LLMs by training small Transformer models on a large dataset of execution traces, including a curriculum of traces from programs with gradually increasing difficulty.

Collectively, these studies show that Transformer-based large language models can model control flow and variable-state dynamics during program execution, demonstrating a capability that strengthens overall code understanding.

Building on these insights, Code World Model (CWM) is the first open-weights LLM that has been trained on Python execution traces during mid-training at a large scale \citep{codgenteam2025_cwm}.
CWM is a \SI{32}{B} parameter dense Transformer LLM that is trained on a large set of Python execution traces stemming from over \SI{120}{M} different functions, \SI{21}{k} executable repository images, and \SI{262}{k} code contest solutions.
The execution traces are formatted as sequences of observation-action pairs conditioned on the code that has been executed. 
The observations correspond to serialized states of program variables, and the actions correspond to the source line being executed, while other information, such as frame event types, is encoded via special tokens.
CWM is able to reliably predict line-by-line execution of Python programs, enabling structured output prediction of programs or grounded reasoning about code generation and execution without access to live execution environments.
While it is possible to build an interactive debugger with CWM by manually steering trace prediction at inference time \citep[Figure B.25]{codgenteam2025_cwm}, it is neither possible to directly jump to future lines of code in constant time nor to predict reverse execution, function inputs, or program termination.

In this work, we introduce \emph{neural debuggers}, which enable all of these capabilities by training language models on execution trace data, where the \emph{next program state is conditioned on debugger actions} such as {breakpoint} or {step into} (see Figure~\ref{fig:state_actions}).

\section{Python program execution traces}
\label{sec:python_exec_traces}
Before Python code is executed by the Python interpreter, it is parsed into an abstract syntax tree and compiled into \emph{code objects}, which contain the operations in the form of bytecode. 
Code objects are created for every Python \emph{code block}, which is executed as a unit, for example, a module, a function body, or a class definition\footnote{\url{https://docs.python.org/3/reference/executionmodel.html}}.
The code objects are executed in Python's \emph{evaluation loop}, which takes the code objects and converts them into a series of stack \emph{frame objects}~\citep{aknin2010pythoninnards, shaw2021cpython}.  
Each frame object contains local and global variables and the code object, which again contains the Python source line and bytecode that is being executed.
We use and record the information at Python's frame object level to create the dataset of execution traces.

Python provides access to runtime events and these stack frame objects by setting a trace function with the signature \texttt{tracefunc(frame,event,arg)} via \texttt{sys.settrace(tracefunc)}\footnote{\url{https://docs.python.org/3/library/sys.html\#sys.settrace}}.
To collect the datasets for our neural debuggers, we use a custom \texttt{tracefunc} to capture execution traces containing \texttt{frame, event} and \texttt{arg}, and execute a large set of Python functions with different inputs as well as repository images with unit tests.
The \texttt{frame} argument is the current stack frame object, containing the local variables and the source line; the \texttt{event} argument is a string describing whether execution enters a new line, is about to enter or return from a function, or if an exception has occurred; and the \texttt{arg} argument contains event-specific data. 
Python debuggers, profilers, and coverage tools use this mechanism to inspect, record, or modify a program's state during execution.
In our case, the recorded execution traces contain the sequences of program states that serve as input data for our neural debugger pipeline, which we describe next.

\section{Neural debugger}

We introduce the concept of \emph{neural debuggers} as neural networks that learn to simulate or predict program execution while providing the core functionalities of conventional debuggers. 
Hence, a neural debugger allows for symbolic debugger-like interactions—such as stepping through execution or inspecting program states—without requiring an executable target program.
We first formalize the notion of a neural debugger by formulating the debugger as a Markov Decision Process (MDP) (Section~\ref{sec:debugger_mdp}). 
Then, we introduce a structured format for training language models on debugger execution trace data (Section~\ref{sec:forward_inverse_trace_format}). 
Finally, we present our practical implementation of the debugger data pipeline, including our action policy for data generation, which enables large-scale training of neural debugger LLMs (Section~\ref{sec:debugger_data_pipeline}).

\subsection{Formulating the debugger as an MDP} %
\label{sec:debugger_mdp}
\begin{figure}[tp!]
    \centering
    \includegraphics[width=0.5\linewidth]{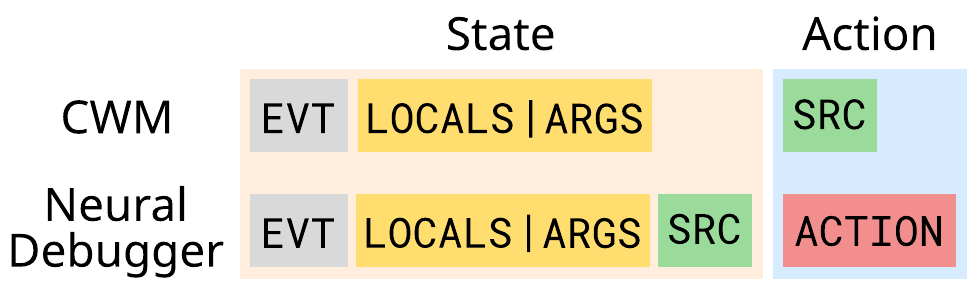}
    \caption{
    State-action structure of Code World Model (CWM) and neural debuggers.
    In CWM~\citep{codgenteam2025_cwm} the actions are viewed as code that modifies the variable states, while in neural debuggers actions influence the program state by controlling program execution analogous to traditional debuggers.
    }
    \label{fig:state_actions}
\end{figure}

A debugger is an interactive software that holds the current state of the program being debugged and receives actions that control program execution, consequently determining the next state of the debugger.
Hence, we model the debugger as an interactive environment, which can be formalized as an MDP given by the tuple $(\gS, \gA, P, R, s_0)$, where
$\gS$ is the space of program states, 
$\gA$ the set of available debugger actions,
$P: \gS \times \gA \rightarrow \gS$ the transition dynamics steered by the debugged program and its input arguments, 
$R: \gS \times \gA \rightarrow \mathbb{R}$ is the reward function, and
$s_0$ the set of possible entry points for the debugger.
In this work, we do not consider a reward function, as we focus solely on state prediction.
Hence, we do not encode an action-prediction task via a reward function, which could be used to specify policies that localize specific program states quickly or mimic the debugging behavior of developers, such as not stepping through all loop iterations.

\paragraph{States.}
The state contains information about the program state and runtime events recorded with \texttt{sys.settrace} (see Section~\ref{sec:python_exec_traces}). 
Specifically, every state contains an event type \texttt{(EVT)}, local variables and their values \texttt{(LOCALS)} or arguments \texttt{(ARGS)}, and the source line of the statement being executed \texttt{(SRC)} (see Figure~\ref{fig:state_actions}).

\vspace{-0.2cm}
\paragraph{Actions.}
The actions (\texttt{ACTION}) of neural debuggers are inspired by the interfaces of traditional debuggers such as \texttt{pdb}\footnote{\url{https://docs.python.org/3/library/pdb.html\#pdbcommand-commands}}.
We categorize them into \emph{step} and \emph{jump} actions: 
\texttt{step\_into} (steps into a function), 
\texttt{step\_over} (jumps over a function call or steps to the next line), 
\texttt{step\_return} (jumps to the return statement of the current function), 
\texttt{breakpoint} (jumps to a specified source line), 
and \texttt{continue} (jumps to the end of the program, i.e., returns the exit code).
Since the concrete action implementations vary between debuggers, we define the outcome of the actions as transitions on the state tree, which we describe next.

\begin{figure}[t!]
    \centering
    \includegraphics[width=\linewidth]{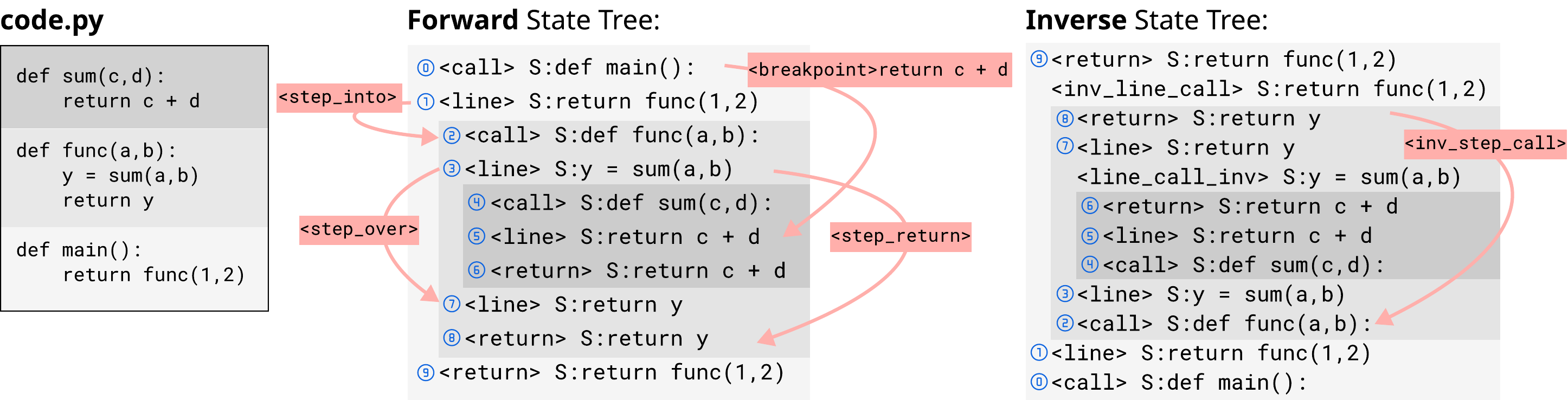}
    \caption{
    Transition model. We visualize the state transitions as traversal on the forward and inverse state tree. \\
    \emph{Left:} Python code. \emph{Middle:} Forward state tree with three levels indicated by indentation. \emph{Right:} Corresponding inverse state tree. The blue numbers illustrate the correspondences between forward and inverse state tree. 
    }
    \label{fig:transition_model}
    \vspace{-0.5cm}
\end{figure}

\vspace{-0.2cm}
\paragraph{State tree.}
At runtime, computer programs maintain a call stack to store information about active subroutines, such as function calls, and to keep track of the point from where the program should continue execution after finishing the execution of a function. 
In Python, the call stack is formed via references between frame objects~\citep[Interpreter Stacks]{aknin2010pythoninnards}.
Since the recorded program state sequences contain information about runtime events (see Section~\ref{sec:python_exec_traces}), the call stack can be reconstructed by keeping track of the order of \texttt{call} and \texttt{return} events.
Specifically, we build a tree data structure that inserts program states belonging to one function call as children of the calling \texttt{line} event node while still retaining the sequential order of the program states. 
In this way, the depth of a node in the tree corresponds to the depth of the call stack at the given program state (see Figure~\ref{fig:transition_model}).

\vspace{-0.2cm}
\paragraph{Transitions.}
Organizing execution traces as state trees provides a foundation for defining the transition model of the debugger as traversal rules on the state tree (see Figure~\ref{fig:transition_model}).
Each debugger action results in a transition from a starting node to a target node on the state tree, defining how the program state evolves under debugger control:
\begin{itemize}[itemsep=1pt, topsep=2pt]
    \item \texttt{step\_into}: The target node is the immediate next node. For starting nodes with a function call, e.g., lines 1 and 3 Figure~\ref{fig:transition_model}, traverse one level deeper in the tree; if at the end of the level, move one level up. Applying only step into actions recovers the recorded program state sequence.
    \item \texttt{step\_over}: The target node is the next node at the current level. If at the end of the level, it moves one level up. It never moves a level down.
    \item \texttt{step\_return}: The target node is the return node with the return event at the current level, e.g., lines 6 and 8 in Figure~\ref{fig:transition_model}.
    \item \texttt{breakpoint SRC}: The target node is the first future node (at any level) that contains the specified source line. If the source line will not be visited in the future, the outcome will be the exit code, i.e., the same as for \texttt{continue}. 
    \item \texttt{continue}: Outcome is the exit code. Exit codes are \texttt{normal} (regular exit), \texttt{error} (an error or uncaught exception occurs), or \texttt{never} (the program enters an infinite loop).
\end{itemize}

\paragraph{Inverse program execution prediction.}
Neural debuggers enable inverse program execution prediction, i.e., inferring plausible predecessor program states, inputs, or function arguments that could have produced a given program state. 
This capability is particularly valuable in automated testing scenarios such as fuzzing, where diverse test inputs must be generated in a semi-random manner.
Unlike reverse debuggers, which allow backward stepping only after a forward execution has been performed and therefore traverse a fixed execution trace~\citep{engblom2012reversedebugging, savidis2021implementationlivereversedebugging}, neural debuggers can start from an arbitrary program state and directly predict plausible predecessors without requiring a prior forward run. 
We note that inverse prediction is inherently ambiguous: program execution is generally many-to-one, making its inverse one-to-many.
For example, reversing a sorting algorithm highlights that many distinct input orderings can yield the same sorted output. 
Even a simple operation such as addition illustrates the issue: given only the sum of two variables, the original operands form an underdetermined system with infinitely many solutions. 
Neural debuggers address this ambiguity by modeling and sampling from the conditional distribution over possible predecessor states, whereas traditional debuggers replay deterministic traces and typically provide neither sampling capabilities nor support for true inverse inference.

\paragraph{Inverse state tree and inverse transitions.}
We construct the inverse program state tree by reversing the order of states in the forward program state tree.
In this process, we duplicate all line event nodes corresponding to function calls and assign them the event type \texttt{inv\_line\_call}.
The program states within the function are then attached as child nodes of these \texttt{inv\_line\_call} nodes, allowing the debugger to either step into or step over a function call depending on whether an \texttt{inv\_step\_into} or \texttt{inv\_step\_over} action is executed (see Figure~\ref{fig:transition_model}).
Finally, breakpoint actions are disabled for inverse program prediction, and the \texttt{step\_return} action is repurposed as \texttt{inv\_step\_call}, which directly predicts a function’s input arguments.

\subsection{Formal language for neural debuggers}
\label{sec:forward_inverse_trace_format}

We introduce a structured language format to represent the state–action sequences generated by both the forward and inverse neural debugger MDPs, designed to ensure compatibility with standard language models.
Specifically, we extend the CWM format~\citep[Section~2.2]{codgenteam2025_cwm}, which includes special separator tokens that mark the beginning of state and action segments, as well as a general mechanism for serializing arbitrary Python objects into text to additionally support debugger actions and inverse execution prediction (see Section~\ref{sec:related_work}~and~Figure~\ref{fig:state_actions}).

\begin{figure}[tp!]
    \centering
    \includegraphics[width=\linewidth]{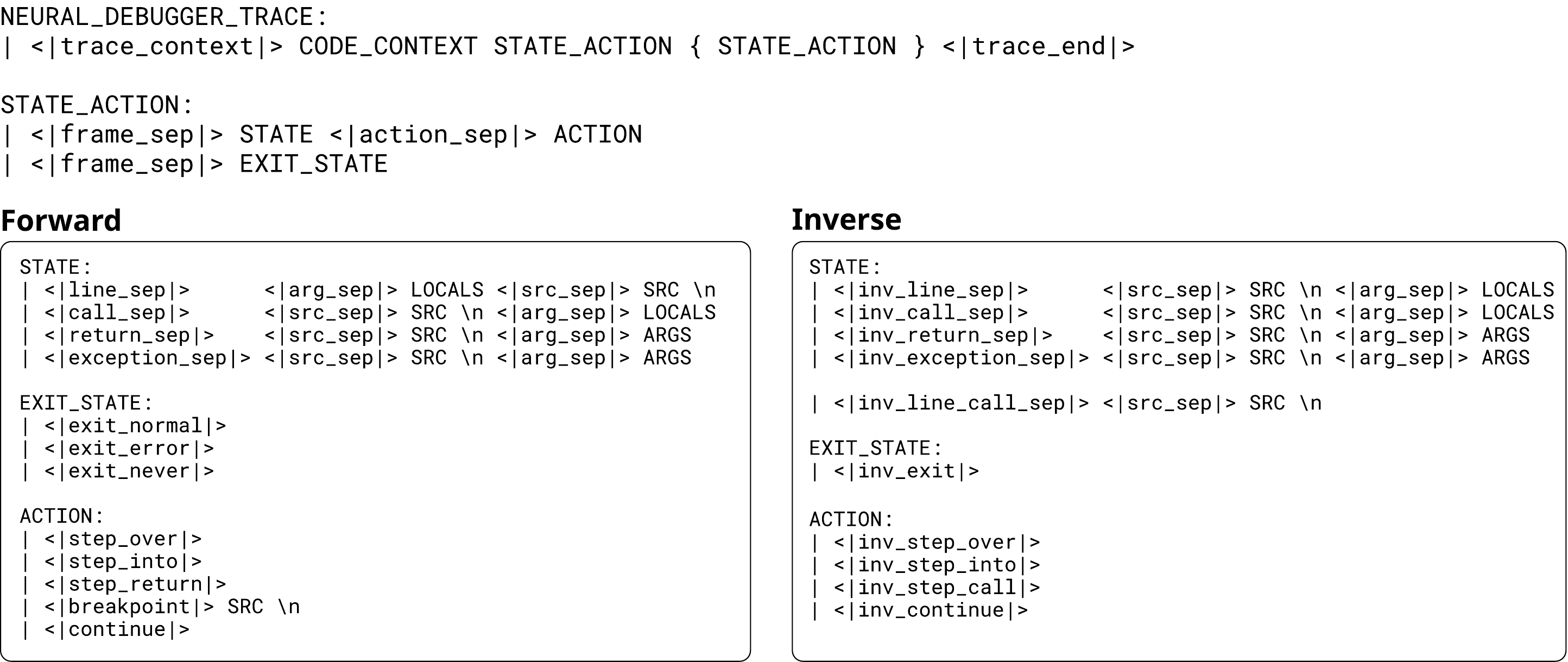}
    \caption{
    Formal neural debugger language grammar. \texttt{|} indicates an OR-statement, \texttt{\{\}} indicate none or more elements, and \texttt{:} denotes an assignment. Whitespaces are shown for illustration purposes only. \texttt{<|.|>} indicate special tokens, \texttt{LOCALS} is the local variable dictionary, \texttt{ARGS} are return or exception arguments, and \texttt{SRC} denotes the source line.
    \label{fig:grammar}}
\end{figure}

\paragraph{Neural debugger language grammar.} 
The grammar of our formal neural debugger language is shown in Figure~\ref{fig:grammar}.
A neural debugger trace consists of a \texttt{CODE} context containing the source code under inspection, followed by a sequence of state–action pairs.
This sequence can represent either a forward or an inverse execution, beginning with an initial state and ending with an \texttt{EXIT\_STATE}.
The boundaries between elements are marked by special separator tokens.
Forward and inverse traces differ in their special event and action tokens, as well as in aspects of the state format.
The detailed structure of states and actions is described in Section~\ref{sec:debugger_mdp}.

\paragraph{Local variable representation.}
The textual representation of local variable dictionaries (\texttt{LOCALS}) must be general enough to serialize arbitrary and potentially large Python objects, while remaining compact to keep token sequence lengths manageable.
Typically, only few local variables change between lines, so we display modified variables only and insert \texttt{"..":".."} to indicate omitted, unchanged entries.
In the forward format, all local variables are shown after scope changes, i.e., following call, return, or breakpoint events.
During inverse prediction, the complete \texttt{LOCALS} dictionary is resolved at call events to ensure prediction of all input variables of the invoked function.
Following~\citet{codgenteam2025_cwm}, \texttt{LOCALS} is serialized as JSON, and arbitrary Python objects are converted to text via their \texttt{\_\_repr\_\_()} methods.

\subsection{Debugger trace data pipeline and dataset}
\label{sec:debugger_data_pipeline}

So far, we have introduced the neural debugger MDP model and a structured text representation for state-action sequences. 
Our full data pipeline produces both forward and inverse trajectories, using a stochastic policy to sample debugger actions. 

\begin{figure}[t!]
    \centering
    \includegraphics[width=\linewidth]{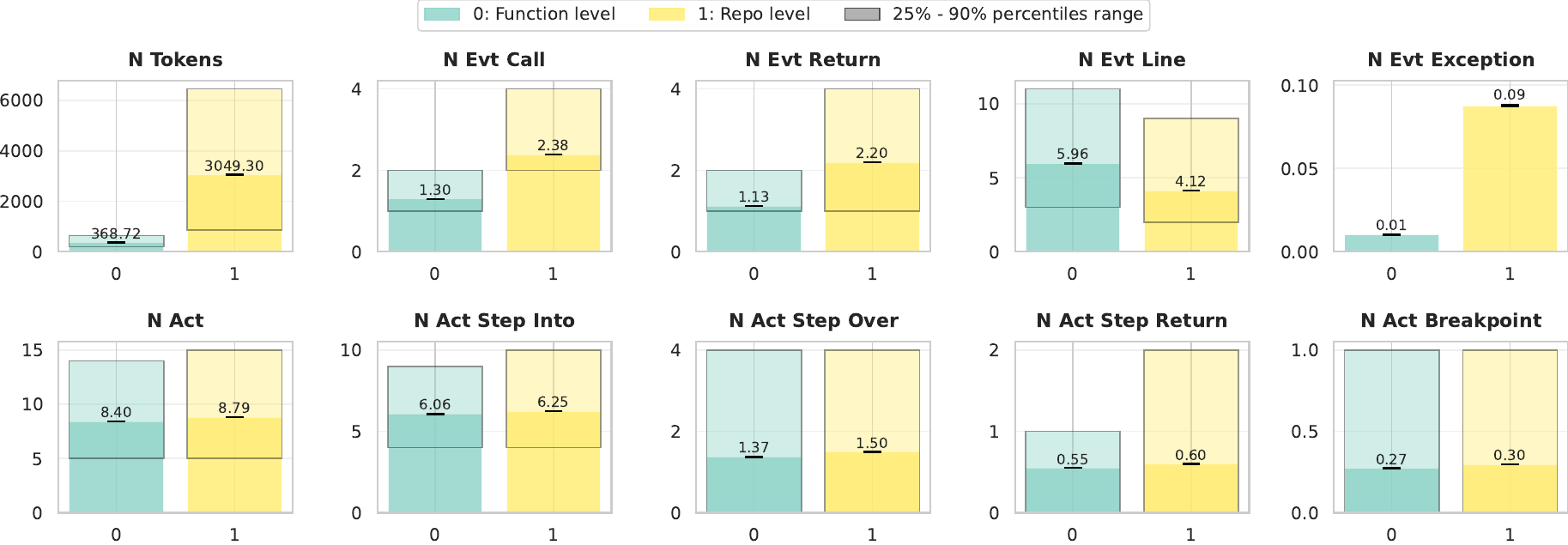}
    \caption{
    Average token, action and event counts of forward debugger trajectory datasets. We show the mean function-level counts in turquoise and the repository-level counts in yellow, with the boxes indicating the 25\% and 90\% range. While the average action counts are similar due to the same action policy, repository-level trajectories contain more function calls, more exceptions, and longer token sequences.
    }
    \label{fig:dataset_statistics_avg_evt_act_counts}
    \vspace{-0.3cm}
\end{figure}

\paragraph{Action policy for data generation.}
To sample debugger state--action trajectories, we define a policy that selects actions conditioned on the current debugger state. 
To ensure broad coverage of available actions and sufficient trajectory lengths, we employ a stochastic policy composed of mixed categorical distributions with carefully chosen probabilities (see Table~\ref{tab:action_policy}). 
Although this setup superficially resembles behavior cloning or imitation learning~\citep{ross2011reductionimitationlearningstructured}, our goal is to model state transitions rather than imitate expert behavior, as we envision coding agents to provide the actions in the future.
As shown in Figure~\ref{fig:dataset_statistics_avg_evt_act_counts}, this random policy produces a diverse set of states and transitions.

\paragraph{Data pipeline.}
Our data pipeline takes as input a sequence of traced program states (see Section~\ref{sec:python_exec_traces}) and the source code blocks containing all source lines visited during tracing.
We then process each execution trace as follows (see Figure~\ref{fig:data_pipeline}):
First, we build the forward or inverse state tree (see Section~\ref{sec:debugger_mdp}).
Second, we sample a debugger state-action trace trajectory from the state tree.
Third, we tokenize the trace trajectory using our structured format defined in Section~\ref{sec:forward_inverse_trace_format}.

\paragraph{Dataset statistics.}
We apply our data pipeline to the function-level and repository-level execution traces from CWM, obtained by running executable Python functions and repository images~\citep[Section~2.2]{codgenteam2025_cwm}.
Because repository-level traces often contain long executions with deep call stacks, we sample a single function call from the stack as the entry point for each debugger trajectory, which truncates the trace to the corresponding function scope.
Using the action policy defined in Table~\ref{tab:action_policy}, we obtain approximately \SI{15}{B} (forward + inverse) repository-level and \SI{100}{B} (forward + inverse) function-level debugger trajectory tokens. %
The stochastic nature of action and entry-point sampling effectively acts as data augmentation, providing sufficient diversity to enable multi-epoch training without overfitting.
We visualize average trajectory statistics from our function- and repository-level datasets in Figure~\ref{fig:dataset_statistics_avg_evt_act_counts}. 
Because both datasets use the same action policy, average action counts (bottom row) are similar.
However, program state event statistics and average token sequence lengths differ (see also Figure~\ref{fig:dataset_statistics_histogram}): repository-level trajectories contain more function calls (i.e., call and return events), more exceptions, and generally longer token sequences for the same number of actions.
This is primarily due to larger local variable dictionaries and the presence of more arbitrary Python objects in repository-level executions.
The following experiments use a mixture of function- and repository-level trajectories in both forward and inverse directions.

\section{Experimental results}
In this section, we explore the feasibility of training neural debuggers for forward and inverse debugging through a systematic empirical evaluation.
Our experiments aim to answer the following questions:
(i) How does finetuning large models that have already been pre-trained on trace data with a different format (e.g., CWM~\citep{codgenteam2025_cwm}) compare to pre-training smaller Transformer models from scratch (Section~\ref{sec:finetuning_pretraining_neural_debuggers})?
(ii) What are the prediction accuracies of individual actions' state elements(Section~\ref{sec:finetuning_pretraining_neural_debuggers}~and~\ref{sec:next_state_element_pred})?
(iii) How well do neural debuggers perform on related downstream tasks such as input and output prediction, and how sensitive is this performance to the prediction horizon (Section~\ref{sec:input_output_cruxeval})? 
\vspace{-0.3cm}
\paragraph{Experiment setup.}
We train neural debugger models by finetuning and pre-training decoder-only Transformer language models with different data mixes and evaluate their forward and inverse program execution prediction capabilities.
Our neural debugger dataset consists of equal proportions of function-level and repository-level execution traces, covering both forward and inverse directions.
We finetune the \SI{32}{B}-parameter CWM model for \SI{50}{B} tokens using a linear warmup followed by a constant learning rate, training exclusively on debugger trace data.
In addition, we pre-train smaller \SI{1.8}{B}-parameter Transformer models on \SI{50}{B} and \SI{150}{B} tokens using a linear warmup and a cosine learning rate decay schedule, and explore three data mixtures: 
debugger trace data only and two different data mixtures of debugger trace data with web data from DCLM~\citep{li2025datacomplmsearchgenerationtraining} and GitHub code data.
Further training details are provided in Appendix~\ref{app:training_recipe}.

\subsection{Finetuning and pre-training neural debuggers}
\label{sec:finetuning_pretraining_neural_debuggers}

\begin{figure}[h!]
    \centering
    \subfloat[Forward prediction.]{
        \includegraphics[width=\linewidth]{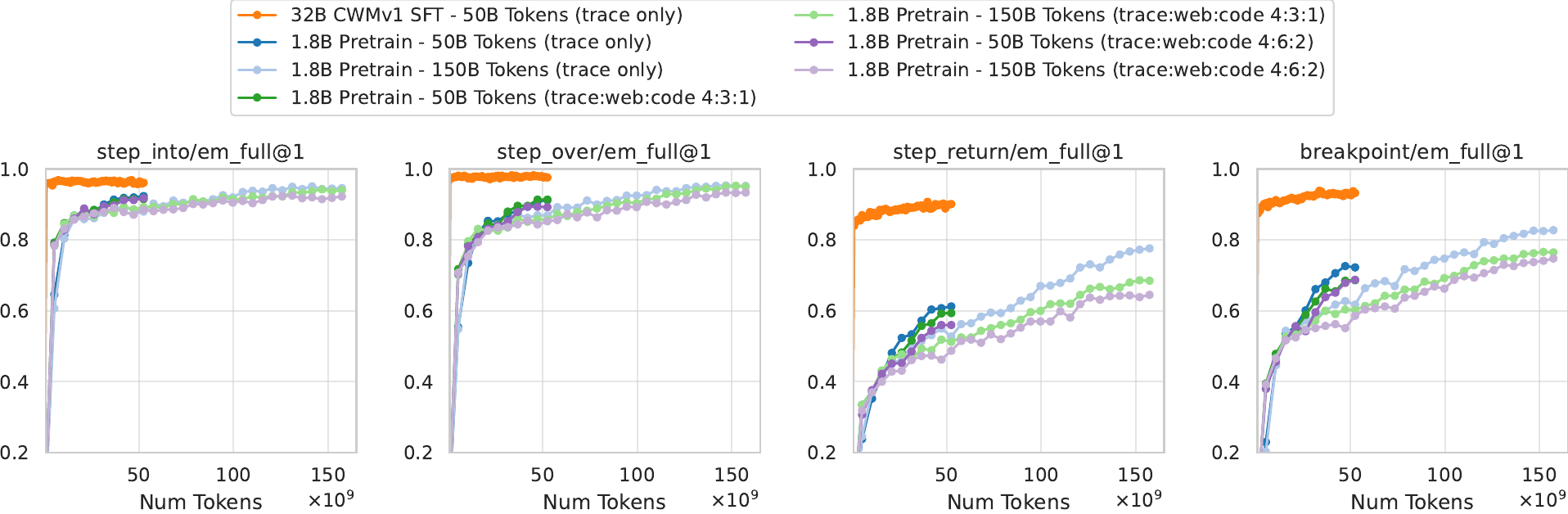}
        \label{fig:train_funclevel_forward}
    }
    \vfill
    \vspace{0.2cm}
    \subfloat[Inverse prediction.]{
        \includegraphics[width=0.75\linewidth]{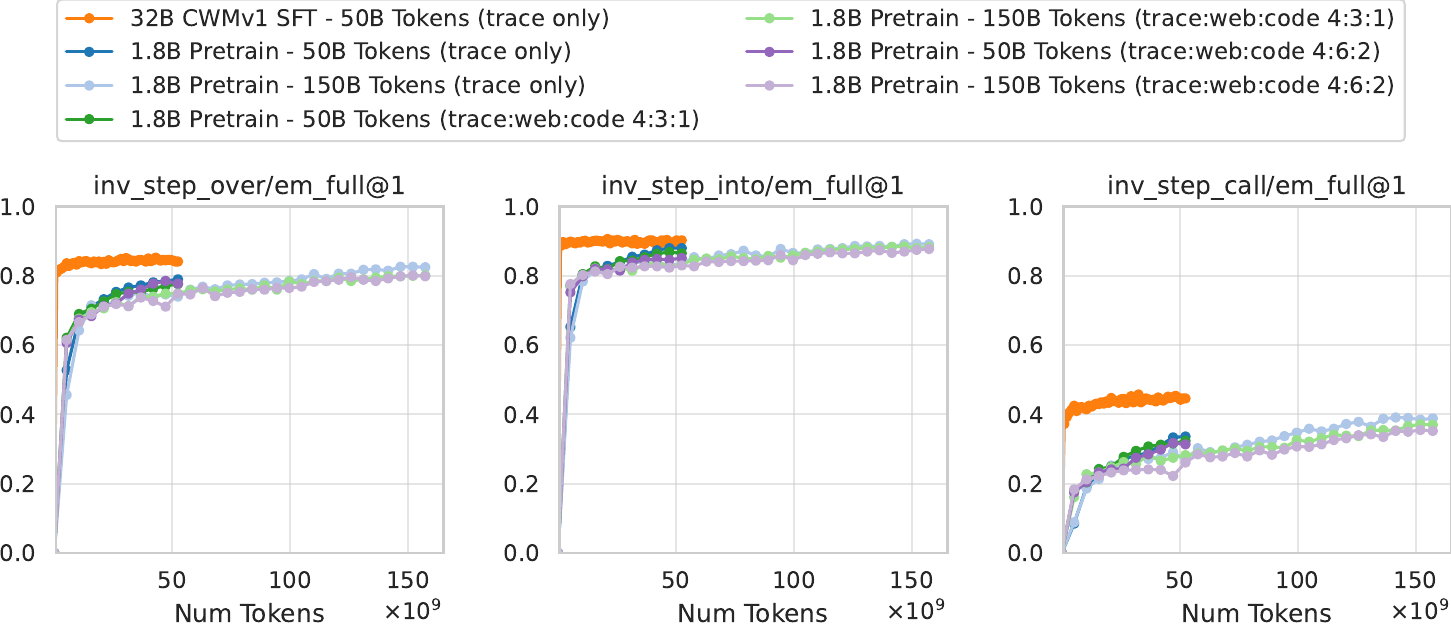}
        \label{fig:train_funclevel_inverse}
    }
    \caption{Evolution of the exact match (em) next state prediction accuracy per debugger action on the \textbf{function-level} validation set during training. (a) Forward execution prediction. (b) Inverse execution prediction. Figure~\ref{fig:train_repolevel} shows the corresponding results for the repository-level validation set, which lead to the same conclusions. \label{fig:train_funclevel}}
    \vspace{-0.2cm}
\end{figure}
\looseness=-1
To compare finetuning and pre-training approaches, we evaluate next-state prediction accuracy per debugger action throughout training.
For each action, we sample 800 trajectories from the validation set using our debugger data pipeline (Section~\ref{sec:debugger_data_pipeline}) and truncate each trace after the respective action to form the evaluation prompt.

In the following, we discuss next-state prediction performance presented in Figure~\ref{fig:train_funclevel} for function-level, and in Figure~\ref{fig:train_repolevel} for repository-level data. 
We report exact-match accuracy between model predictions (with greedy decoding) and ground-truth program states. 
Our results for function-level and repository-level evaluations lead to the same conclusions, which is why we place Figure~\ref{fig:train_repolevel} in the appendix.
\vspace{-0.2cm}
\paragraph{Step actions are easier than jump actions.}
By comparing the left two columns to the right two columns in Figure~\ref{fig:train_funclevel_forward}~and~\ref{fig:train_repolevel_forward}, we observe that step actions (e.g., step into, step over) achieve higher exact-match accuracy than jump actions (e.g., step return, breakpoint).
While step actions plateau early, jump actions continue to improve with more training tokens.
This difference arises because predicting transitions to the next source line is simpler than predicting multi-line transitions, and single-line transitions occur far more frequently in the dataset (Figure~\ref{fig:dataset_statistics_avg_evt_act_counts}).
In addition, in Figure~\ref{fig:train_funclevel}~and~\ref{fig:train_repolevel} the finetuned CWM model shows a rapid performance increase over the first few tokens, indicating a benefit from its pre-training and prior exposure to trace data during its mid-training phase, even though the trace data format differed and did not include debugger actions~\citep{codgenteam2025_cwm}.
\vspace{-0.2cm}
\looseness=-1
\paragraph{Inverse execution is learnable.} %
In Figure~\ref{fig:train_funclevel_inverse}~and~\ref{fig:train_repolevel_inverse}, we observe a similar trend for inverse execution prediction as for forward execution prediction in Figure~\ref{fig:train_funclevel_forward}~and~\ref{fig:train_repolevel_forward}.
Even though inverse prediction accuracies are lower than forward accuracies, they consistently improve over training. 
We note that for the \texttt{inv\_step\_call} action in the right column of Figure~\ref{fig:train_funclevel_inverse}~and~\ref{fig:train_repolevel_inverse}, only the trend of the exact match @1 metric is indicative, not the absolute numbers, since the exact match @1 metric does not capture the inherent ambiguity in the inverse program states.
Since forward prediction is deterministic (in most cases), perfect accuracy is achievable. However, for inverse prediction, the inherent ambiguity provides a data-dependent upper bound on performance.
In contrast, in Table~\ref{tab:cruxeval_input_output}, we compute the pass@1 metric with \texttt{assert f(predicted\_input) == reference\_output}; i.e., accounting for the ambiguity, we observe that input prediction accuracies are comparable to the output prediction. 
We refer to Appendix~\ref{app:eval_inverse_prediction} for details on the evaluation of inverse execution prediction.
\vspace{-0.2cm}
\looseness=-1
\paragraph{Small models are good neural debuggers.}
Comparing the finetuned \SI{32}{B}-parameter CWM model with our \SI{1.8}{B}-parameter models trained on the same \SI{50}{B} tokens in Figure~\ref{fig:train_funclevel_forward}~and~\ref{fig:train_repolevel_forward}, 
we observe a modest $\sim$\SI{5}{\percent} point gap for step actions and a larger $>$\SI{15}{\percent} point gap for more complicated jump actions.
However, extending pre-training to \SI{150}{B} tokens substantially narrows both gaps, suggesting that small Transformers can already serve as capable neural debuggers.
Furthermore, experiments with different data mixtures indicate that debugger data can be integrated into existing pre- or mid-training corpora, similar to CWM’s mid-training strategy~\citep{codgenteam2025_cwm}.

\subsection{Next program state prediction by state component}
\label{sec:next_state_element_pred}

For more fine-grained performance analysis, we evaluate next-state prediction accuracies by debugger action and state element (Figure~\ref{fig:state_actions}).
Here, we train all models exclusively on debugger trace tokens and report accuracies for four components: local variable dictionaries, return or exception arguments, source lines, and state events.

\begin{figure}[t!]
    \centering
    \subfloat[Forward prediction.]{
        \includegraphics[width=0.8\linewidth]{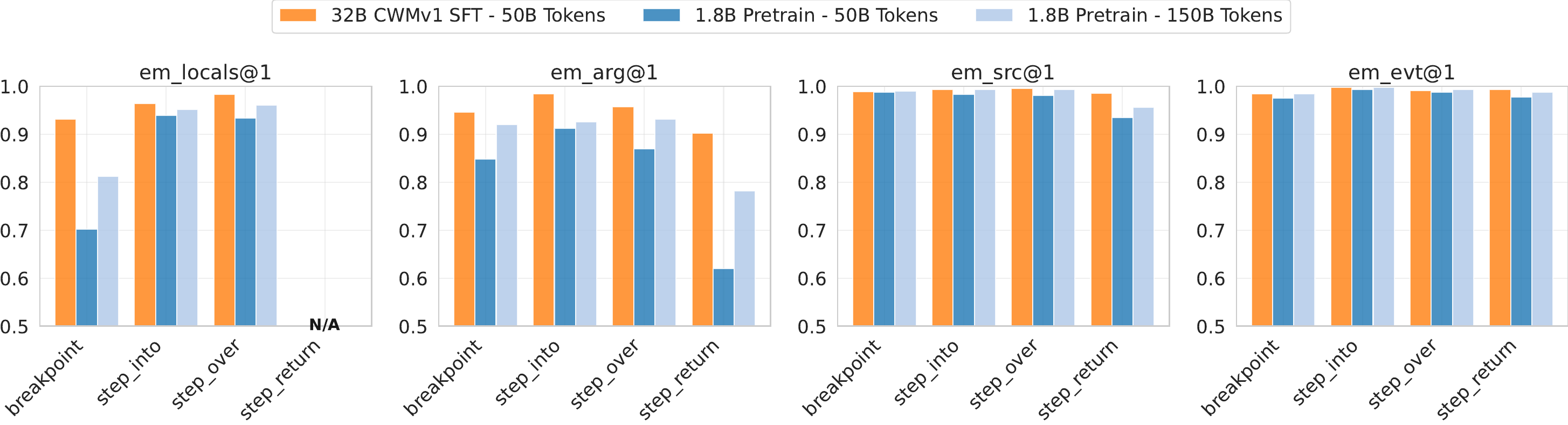}
        \label{fig:next_step_component_funclevel_forward}
    }
    \vfill
    \vspace{0.2cm}
    \subfloat[Inverse prediction.]{
        \includegraphics[width=0.8\linewidth]{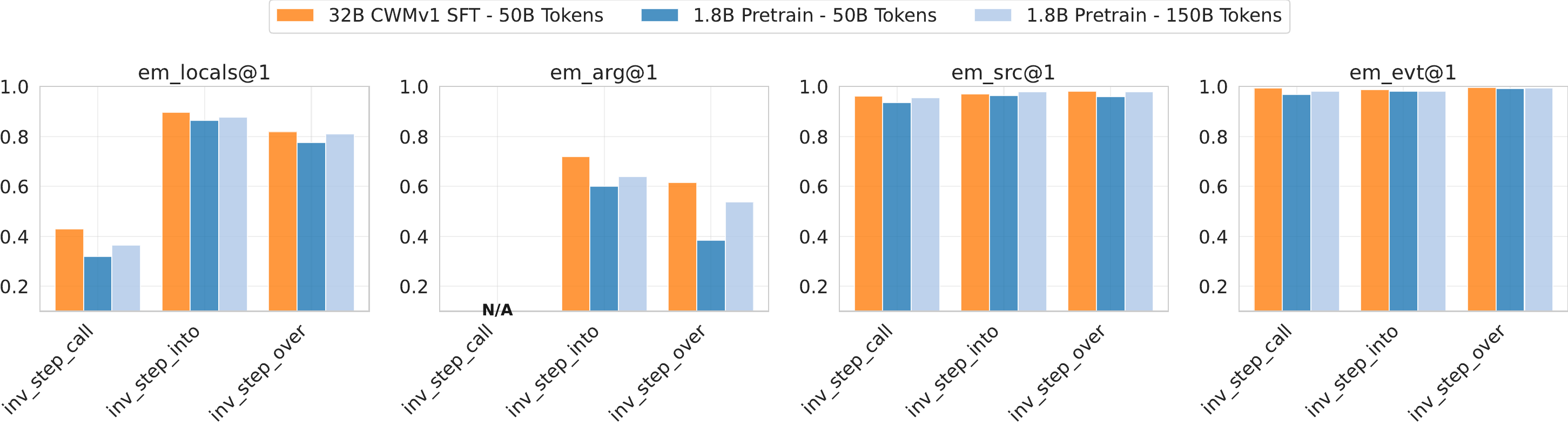}
        \label{fig:next_step_component_funclevel_inverse}
    }
    \caption{Exact match (em) next state prediction by state component per debugger action on the \textbf{function-level} validation set. Most prediction errors are in the local variables and return or exception arguments, while source lines and state events are predicted reliably. Figure~\ref{fig:next_step_component_repolevel} shows the corresponding results for the repository-level data. Conclusions are the same. \label{fig:next_step_component_funclevel}}
\end{figure}

\paragraph{Source lines \& state events are predicted reliably; local variables contain errors.}
Across all actions and datasets, our models consistently achieve high accuracies for source line and state event prediction in both forward and inverse modes for function-level (see Figure~\ref{fig:next_step_component_funclevel}) and repository-level (see Figure~\ref{fig:next_step_component_repolevel}) data.
In addition, we find that predicting source lines for function-level data achieves higher exact match scores than for repository-level data, especially for \texttt{step\_return} action prediction of the \SI{1.8}{B} models, which exhibits a $\sim$\SI{5}{\percent} point (function-level) and $\sim$\SI{10}{\percent} point (repository-level) drop in source line accuracy (see em\_src@1 in Figure~\ref{fig:next_step_component_funclevel_forward}~and~\ref{fig:next_step_component_repolevel_forward}).
We hypothesize that this is due to more complex conditional branching in repository-level data.
Most remaining errors stem from local variable and return/exception argument predictions, which show markedly lower accuracies than source line and state event predictions.
We further find that the accuracy gaps between the finetuned CWM model and smaller pre-trained models---particularly for jump actions such as breakpoint or \texttt{step\_return}---originate primarily from these local variables and argument components (e.g., em\_locals, em\_arg) instead of source line and event components (e.g., em\_src, em\_evt).
This indicates that predicting variables values is harder than predicting control flow, especially for smaller models.
Note that, similar to the previous section, em\_locals@1 does for inv\_step\_call actions in Figure~\ref{fig:next_step_component_funclevel_inverse}~and~\ref{fig:next_step_component_repolevel_inverse}, it does not account for ambiguities in inverse execution prediction (see Section~\ref{app:eval_inverse_prediction}).
Entries marked “N/A” indicate state components not present in a given trace element (see Figure~\ref{fig:grammar}).

\subsection{Input and output prediction on CruxEval}
\label{sec:input_output_cruxeval}

To compare the code understanding and execution capabilities of neural debuggers with general-purpose language models, we evaluate them on the CruxEval input–output prediction benchmark~\citep{gu2024cruxeval}.
We generate debugger execution traces from CruxEval's Python functions and define prediction tasks corresponding to debugger jump actions: \texttt{step\_return} and \texttt{breakpoint} for output prediction (Figure~\ref{fig:cruxeval_output_prompts}), and \texttt{inv\_step\_call} for input prediction (Figure~\ref{fig:cruxeval_input_prompt}).
For each case, we construct custom prompts such that the target state after the respective jump action contains either the output value or the input arguments of the function.

\paragraph{Neural debuggers excel at output prediction.}
Table~\ref{tab:cruxeval_input_output} summarizes CruxEval input and output prediction pass@1 scores calculated with \texttt{assert f(input) == output} for our models.
We observe consistently high input and output prediction performance, with \texttt{breakpoint} outperforming \texttt{step\_return} for output prediction.
We attribute this to the prompt design: the \texttt{breakpoint} action explicitly includes the source line associated with the return statement, helping the model localize the correct execution context and focus on predicting the function’s output rather than identifying the relevant line.
Our neural debugger model finetuned from CWM achieves a CruxEval score of \SI{83.2}{\percent} using the \texttt{breakpoint} action.
Even without providing the return line in the prompt with the \texttt{step\_return} action, our neural debugger model finetuned from CWM achieves a CruxEval score of \SI{77.9}{\percent}, corresponding to a \SI{19.8}{\percent}-point improvement compared to the stock CWM model evaluated with the CWM execution trace format~\citep[Table 8, Trace Step, \SI{58.1}{\percent}]{codgenteam2025_cwm}.
Already, the smaller \SI{1.8}{B} Transformer trained from scratch achieves \SI{57.7}{\percent} and \SI{48.0}{\percent} with \texttt{breakpoint} and \texttt{step\_return} action on the same task after \SI{150}{B} training tokens, highlighting that training exclusively on debugger trace data confers strong execution reasoning abilities.

\begin{table}[t!]
\centering
\begin{tabular}{cc|c|cc}
\toprule
\multirow{2}{*}{Model} &\multirow{2}{*}{Tokens}  & Input & \multicolumn{2}{c}{Output} \\
&& \texttt{inv\_step\_call} & \texttt{step\_return} & \texttt{breakpoint} \\ 
\midrule
1.8B Transformer Pretrain & 50B & 40.7 & 34.4 & 44.9 \\
1.8B Transformer Pretrain & \SI{150}{B} & 53.6 & 48.0 & 57.7 \\
\midrule
32B CWM Finetune & 50B & 66.5 & 77.9 & 83.2 \\
\bottomrule
\end{tabular}
\caption{CruxEval input and output pass@1 scores for single step prediction with neural debugger actions \texttt{inv\_step\_call}, \texttt{step\_return}, and \texttt{breakpoint} (greedy decoding). All models are trained on debugger trace data only.}
\label{tab:cruxeval_input_output}
\end{table}

\begin{figure}[h!]
    \centering
    \subfloat[Input prediction with inverse \texttt{inv\_step\_call} action (exact match of local variable dict).]{
        \includegraphics[width=0.75\linewidth]{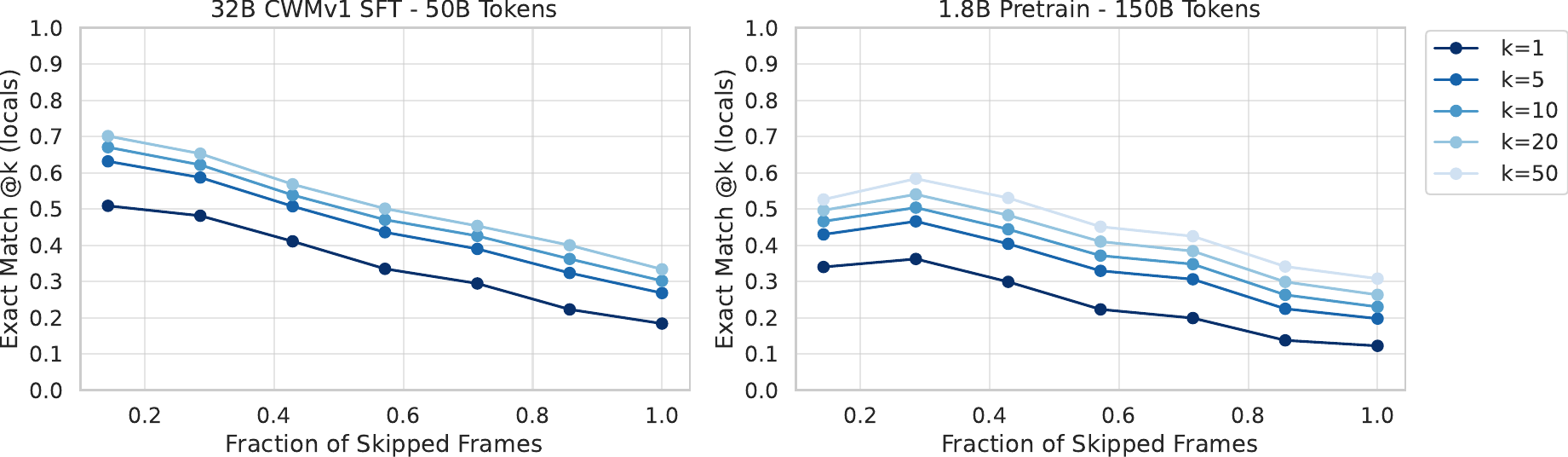}
        \label{fig:cruxeval_input_step_call}
    }
    \vfill
    \vspace{0.2cm}
    \subfloat[Output prediction with forward \texttt{step\_return} action (exact match of the return argument).]{
        \includegraphics[width=0.75\linewidth]{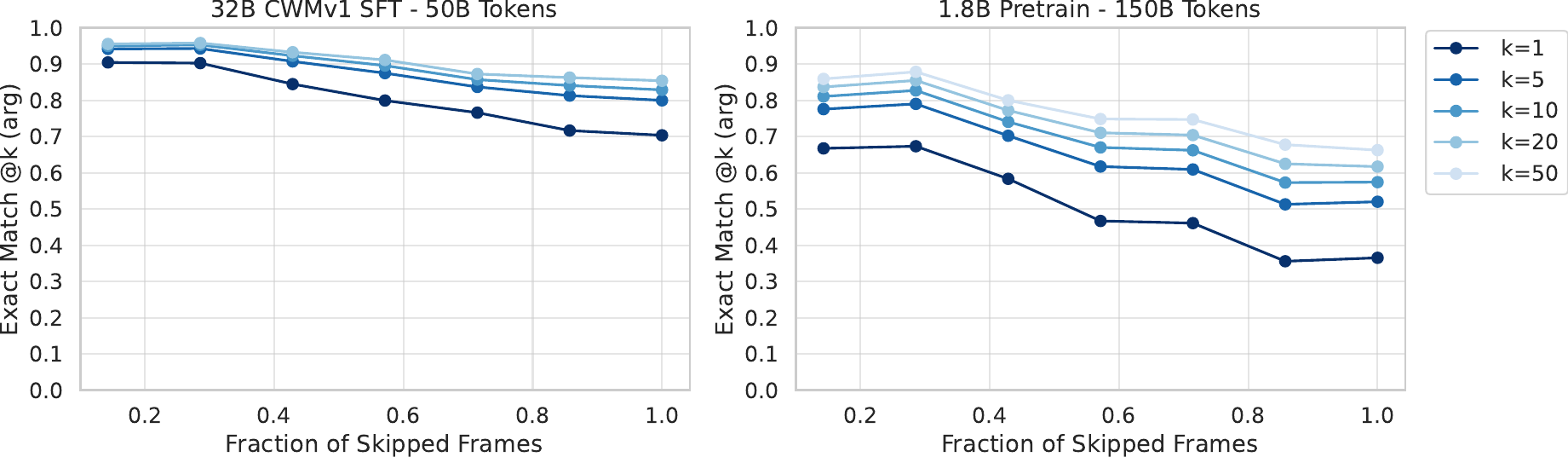}
        \label{fig:cruxeval_output_step_return}
    }
    \vfill
    \vspace{0.2cm}
    \subfloat[Output prediction with forward \texttt{breakpoint} action (exact match of local variable dict).]{
        \includegraphics[width=0.75\linewidth]{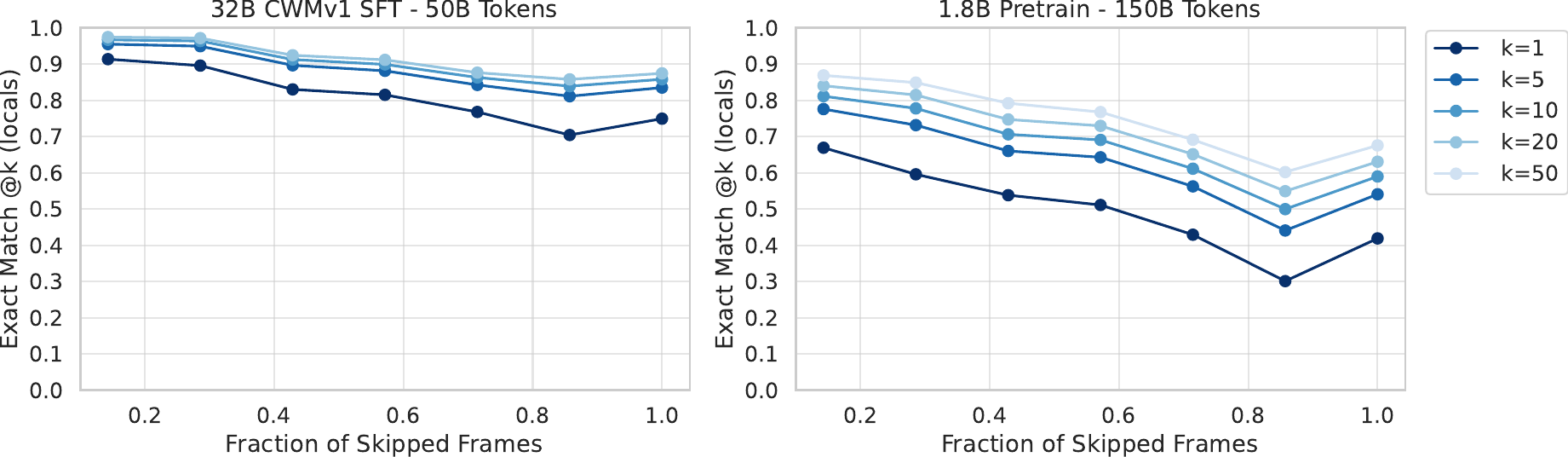}
        \label{fig:cruxeval_output_breakpoint}
    }
    \caption{CruxEval input and output exact match @k scores for increasing prediction horizon (see Section~\ref{app:eval_inverse_prediction} for a discussion on exact~match~vs.~pass metric). We use the same prompts (see Section~\ref{app:cruxeval_in_out_prompts}) as for Table~\ref{tab:cruxeval_input_output} and insert step actions to increase number of revealed program states in the prompt. Long prediction horizons correspond to a high fraction of skipped frames (close to 1) and short prediction horizons to a low fraction (close to 0). We generate responses with temperature 0.6 and top-p 0.95. Across all models, the prediction accuracy decreases as the prediction horizon increases,
with a steeper decline for smaller models trained from scratch.. \label{fig:cruxeval_inout_main}}
    \vspace{-0.0cm}
\end{figure}

\paragraph{Prediction accuracy decreases with prediction horizon.}
We next analyze how input and output prediction accuracy varies with the prediction horizon.
For each CruxEval Python function, we generate multiple prompts by inserting a varying number of \texttt{step\_into}, \texttt{step\_over}, or \texttt{inv\_step\_over} actions before the final jump action (\texttt{breakpoint}, \texttt{step\_return}, or \texttt{inv\_step\_over}).
This setup forces the final jump action to skip a variable number of intermediate program states, ranging from the next state (single-step prediction) to direct prediction of the full function’s input or output.
We normalize the number of skipped states by the total number of states in the function and group examples with similar normalized horizons into bins.
Figure~\ref{fig:cruxeval_inout_main} plots the fraction of skipped states on the x-axis (0 corresponding to single-step prediction and 1 corresponding to full input/output prediction as in Table~\ref{tab:cruxeval_input_output}), and plots input and output prediction accuracy on the y-axis for different exact match @k thresholds (analogous to pass@k, except for input prediction in Figure~\ref{fig:cruxeval_input_step_call}; see Section~\ref{app:eval_inverse_prediction} for details).
For both models, the finetuned \SI{32}{B} CWM neural debugger and the \SI{1.8}{B} model trained from scratch, accuracy decreases as the prediction horizon increases, with a steeper decline for smaller models trained from scratch.
Larger sampling budgets (higher $k$) partially mitigate the accuracy drop, suggesting that test-time ensembling, majority voting, or resampling based on model uncertainty could serve as effective strategies to further improve neural debugger performance.
We believe this result offers exciting avenues for future research: for example, one could investigate whether model uncertainties are suitable for deciding when to jump to return events, adaptively assigning inference compute based on program difficulty.

\newpage
\section{Limitations and future work}
\label{sec:limitations_future_work}

\looseness=-1
Neural debuggers represent a step towards program execution-aware language models with promising practical applications.
At this stage, however, several limitations remain that suggest clear directions for further research.

\paragraph{Agentic program repair, reasoning \& tool use with neural debuggers.}
As a first downstream application, we evaluated neural debuggers on input and output prediction. 
We believe that including neural debuggers in agentic coding and extending them to tasks such as program repair and bug fixing offer particularly promising opportunities.
Such applications could benefit from LLMs that self-debug their generated code during reasoning or from controlling a debugger in real debugging environments.

\paragraph{Expanding \& improving data generation.}
Thus far, we have applied neural debuggers exclusively to Python programs and used random action policies to generate debugger trajectories. 
Future work could expand the dataset with execution traces from additional programming languages and develop more structured or goal-directed action policies.
For example, incorporating syntactic code information—such as compound statements\footnote{\url{https://docs.python.org/3/reference/compound_stmts.html}}, e.g., conditions or loops—into the data generating policy could further bias trajectories toward semantically richer transitions and improve data quality.
As a result, we would expect different action distributions for different code data sources (e.g., for function-level and repository-level data in Figure~\ref{fig:dataset_statistics_avg_evt_act_counts}).

\paragraph{Improving inverse debugging.}
Further research on neural debuggers could target improvements in local variable prediction through better modeling of ambiguity and feasible value sets.
Additionally, evaluation metrics for inverse prediction should be adapted to account for multiple valid traces or outcomes. 
Such metrics could capture more nuanced prediction errors of neural debuggers, especially for inverse debugging, which would naturally extend our analysis in Section~\ref{sec:next_state_element_pred}~and~\ref{sec:input_output_cruxeval}.

\paragraph{Better Python object representations.}
Our current approach serializes arbitrary Python objects to text using built-in mechanisms.
While this simplifies data collection, it becomes infeasible for very large or complex data structures. 
We observed this in the longer state-action trajectories for repository-level data (see Figure~\ref{fig:dataset_statistics_histogram}).
Developing compact neural representations of arbitrary Python objects remains an interesting problem for future research.

\section{Conclusion}
In this work, we introduced the concept of \emph{neural debuggers}—neural networks capable of predicting the line-by-line execution of computer programs conditioned on common debugger actions such as \texttt{step\_into}, \texttt{step\_over}, \texttt{breakpoint}, and \texttt{step\_return}.
We formalized the neural debugger as a Markov Decision Process (MDP), where states comprise program variables and source lines, and transitions are defined by traversal rules on a tree structure reconstructed from execution traces via the program’s call stack.
Our experiments show that finetuning existing large language models or pre-/mid-training on debugger trace data yields neural debuggers that achieve accurate intermediate state predictions and strong overall execution prediction performance.
We believe that neural debuggers represent a promising step toward language models that are explicitly grounded in program execution. 
By learning to model step-wise execution dynamics and debugger control flow within a neural framework, they open a path toward integrating reasoning and execution within a single learned system. 
Looking forward, we envision neural debuggers as a core component of future agentic coding systems, acting as a world model for simulated debugging environments and enabling agents to interact with real debuggers through execution-aware feedback. 
In this way, neural debuggers tightly couple neural reasoning with executable program behavior and have the potential to substantially advance code generation, understanding, and debugging.

\clearpage 

\bibliographystyle{assets/plainnat}
\bibliography{main}

\newpage
\appendix
\beginappendix

\counterwithin{table}{section}
\counterwithin{figure}{section}

\section{Extended neural debugger}
\label{app:ext_neural_debugger}

\subsection{Extended debugger trace dataset}
\label{app:ext_debugger_trace_dataset}

\begin{figure}[h!]
    \centering
    \includegraphics[width=0.8\linewidth]{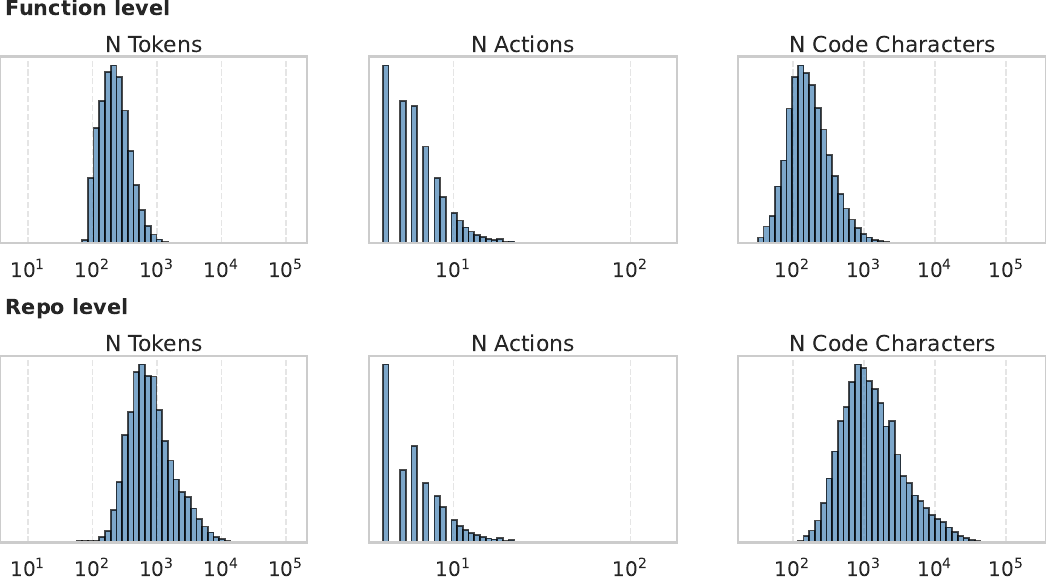}
    \caption{
    Length distributions of the debugger trace dataset. We show the length distribution histograms of debugger trajectories in number of tokens and actions, and show the distribution of the code length in number of characters.
    Even though the number of actions are similar for function-level and repository-level data, the average token count for repository-level trajectories is significantly higher (see also Section~\ref{sec:debugger_data_pipeline}). The reasons for the longer trajectories are more arbitrary python objects in their local variable state dictionaries and larger code context (see right column).  
    }
    \label{fig:dataset_statistics_histogram}
\end{figure}

\paragraph{Action probabilities for data generation (Table~\ref{tab:action_policy}).}
We generate our debugger trace trajectory dataset by sampling actions from a mixture of two categorical distributions, each selected with equal probability (see Table~\ref{tab:action_policy}).
The first distribution samples uniformly from the full set of available actions to ensure that every action type is represented in the trajectories.
However, actions such as breakpoint, step return, or continue can skip over large portions of the program’s execution, resulting in shorter trajectories.
To mitigate this effect, we introduce a second distribution that samples exclusively from the step into and step over actions, with equal probability between them.
This balanced mixture encourages sufficient trajectory length while maintaining diversity in the action space.

\begin{table}[h!]
\centering
\begin{tabular}{c|cccccc}
\toprule
\text{Policy Prob} & \texttt{step\_into} & \texttt{step\_over} & \texttt{step\_return} & \texttt{breakpoint} & \texttt{continue}  \\ 
\midrule
0.5 & 0.35 & 0.1 & 0.2 & 0.1 & 0.05 \\
0.5 & 0.5 & 0.5 & - & - & - \\ 
\bottomrule
\end{tabular}
\caption{Action probabilities for the action policy mix used to generate our debugger trace dataset.}
\label{tab:action_policy}
\end{table}

\subsection{Evaluating Inverse Execution Prediction}
\label{app:eval_inverse_prediction}
The standard way to evaluate input and output predictions is to execute the code and check its correctness with \texttt{assert} statements and given reference inputs and outputs~\citep{gu2024cruxeval}.
For input prediction, we use \texttt{assert(predicted\_input) == reference\_output}, while for output prediction, we use \texttt{assert(reference\_input) == predicted\_output}.
In order to evaluate our neural debuggers on CruxEval, we collect ground truth execution traces for all examples with the provided reference inputs. 
We then use these traces to create prompts and targets for the evaluations in Table~\ref{tab:cruxeval_input_output} and Figure~\ref{fig:cruxeval_inout_main}. 
Similarly, we use the recorded execution traces (see Section~\ref{sec:debugger_data_pipeline}) from our validation sets to generate prompts and targets from these execution trace validation sets for Figure~\ref{fig:train_funclevel} and Figure~\ref{fig:next_step_component_funclevel}.
To compute the metrics, we first parse the predictions and then compare the individual state components (see Figure~\ref{fig:state_actions}) to the ground truth via exact match.

For forward debugger trace prediction, this approach corresponds to the metric computation in CruxEval, as most functions and repositories in our validation sets and CruxEval functions are deterministic. 
In fact, CruxEval even applies a filter to contain only deterministic functions\footnote{\url{https://github.com/facebookresearch/cruxeval/blob/main/data/README.md}}.
In this case, exact\_match@1 and pass@1 metric scores are identical. 

However, programs that are deterministic in forward execution must not be deterministic in inverse execution; e.g., remember our previous example of the inverse execution of a sum operation. 
Similarly, our example in Figure~\ref{fig:debugger_trace_example_count_r} illustrates this fact.
Therefore, using only a single reference input candidate to compute exact match scores for predicted function arguments can lead to an underestimation of performance, as we show in Table~\ref{tab:cruxeval_input_em_vs_pass_at}.
For the interpretation of our results, this means that our exact match scores still provide a signal about the relative performance between models, but they may underestimate the true absolute performance.
To address this problem, we plan to develop dedicated evaluations and benchmarks focused on neural debugger capabilities, using executable functions and/or Docker images to produce target traces based on data (such as inputs or actions) generated by the model.

\begin{table}[t!]
\centering
\begin{tabular}{cc|c|c}
\toprule
\multirow{2}{*}{Model} &\multirow{2}{*}{Tokens}  & Input & Input \\
&& \texttt{inv\_step\_call} & \texttt{inv\_step\_call} \\
&& exact\_match@1 & pass@1 \\
\midrule
1.8B Transformer Pretrain & 50B & 14.3 & 40.7 \\
1.8B Transformer Pretrain & \SI{150}{B} & 17.7 & 53.6 \\
\midrule
32B CWM Finetune & 50B & 23.1 & 66.5 \\
\bottomrule
\end{tabular}
\caption{Comparison of CruxEval input exact\_match@1 and pass@1 scores for single step prediction with the neural debugger action \texttt{inv\_step\_call} (greedy decoding) corresponding to Table~\ref{tab:cruxeval_input_output}. The exact\_match@1 (em@1) scores are considerably lower than pass@1 score, since exact match only compares to the reference inputs, and does not account for potential ambiguities in input predictions.}
\label{tab:cruxeval_input_em_vs_pass_at}
\end{table}

\section{Extended experiments}
\label{app:ext_experiments}

\subsection{Training recipe}
\label{app:training_recipe}
We train our neural debuggers using the AdamW optimizer~\citep{loshchilov2018decoupled} with weight decay \SI{0.1}{}. 
We use the Llama-2 architecture for the pre-training of our neural debugger models with \SI{1.8}{B} parameters~\citep{touvron2023llama2openfoundation}.
For pre-training we use a learning rate schedule consisting of \SI{750}{} linear warmup steps to the peak learning rate of \SI{1e-3}{}, followed by a cosine decay to learning rate zero, over the remaining training steps. 
In contrast, for finetuning, we use a linear warmup over \SI{750}{} steps, followed by a constant peak learning rate of \SI{1e-5}{}. 
For all experiments we use a sequence length of \SI{16384}{} and a batch size of \SI{1}{M} tokens or \SI{64}{} sequences.

\subsection{Extended next state prediction results}

In the main text in Section~\ref{sec:finetuning_pretraining_neural_debuggers} and Section~\ref{sec:next_state_element_pred}, we show the results for the function-level data.
In addition, we report the analogous results for the repository-level data.

In Figure~\ref{fig:train_repolevel}, we show the evolution of the next state prediction accuracy per debugger action on the repository-level validation set during training. 

\begin{figure}[h!]
    \centering
    \subfloat[Forward prediction.]{
        \includegraphics[width=\linewidth]{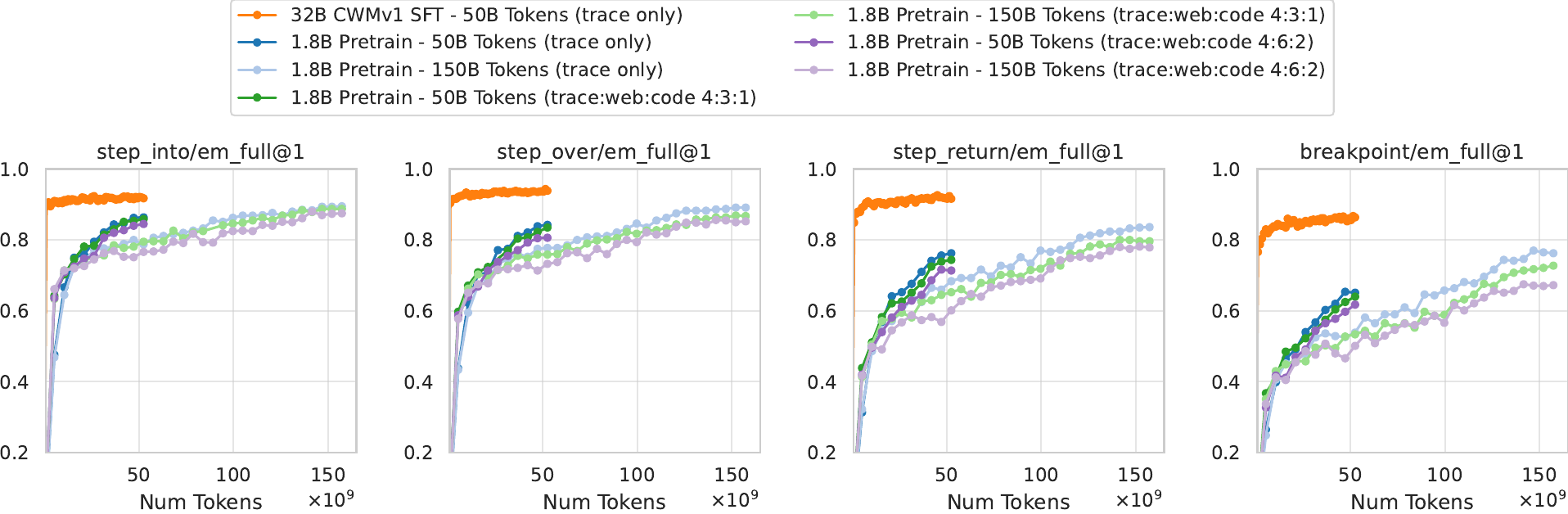}
        \label{fig:train_repolevel_forward}
    }
    \hfill
    \subfloat[Inverse prediction.]{
        \includegraphics[width=0.75\linewidth]{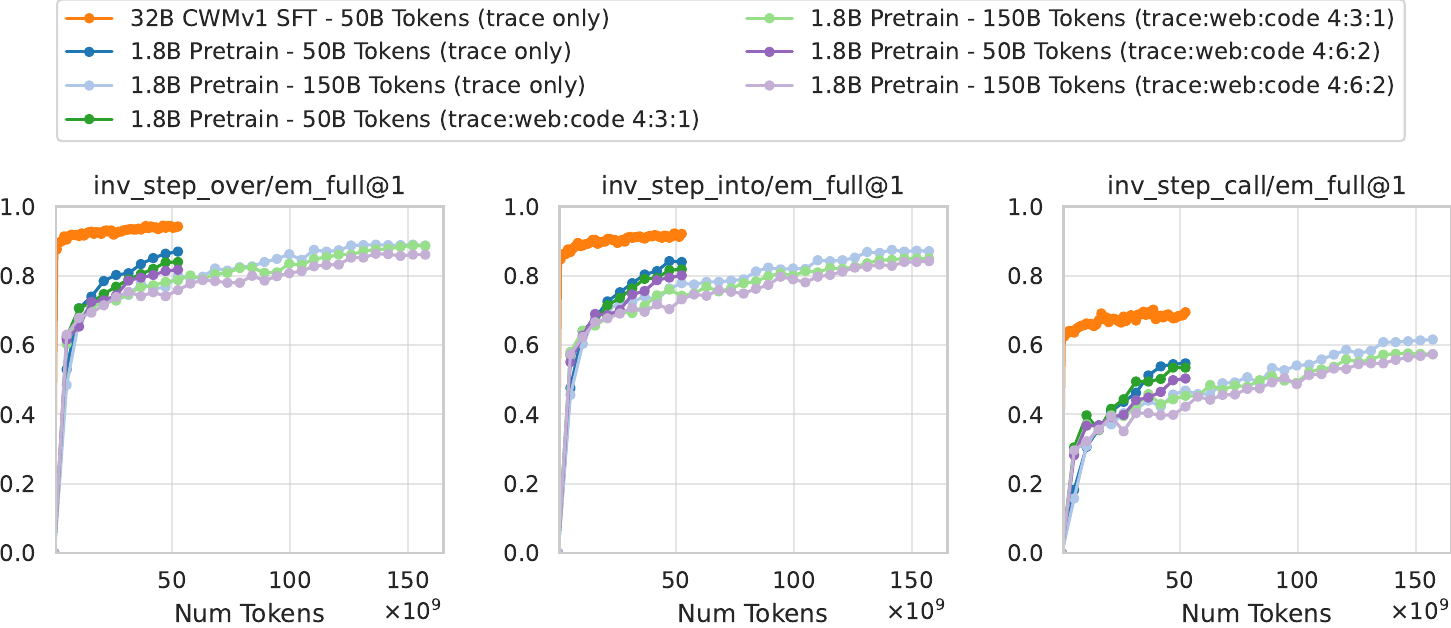}
        \label{fig:train_repolevel_inverse}
    }
    \caption{Evolution of the exact match (em) next state prediction accuracy per debugger action on the \textbf{repository-level} validation set during training. (a) Forward execution prediction. (b) Inverse execution prediction. Figure~\ref{fig:train_funclevel} shows the corresponding results for the function-level validation set. The results on both validation sets lead to similar conclusions. \label{fig:train_repolevel} \label{fig:train_repolevel}}
\end{figure}

\begin{figure}[h!]
    \centering
    \subfloat[Forward prediction.]{
        \includegraphics[width=0.8\linewidth]{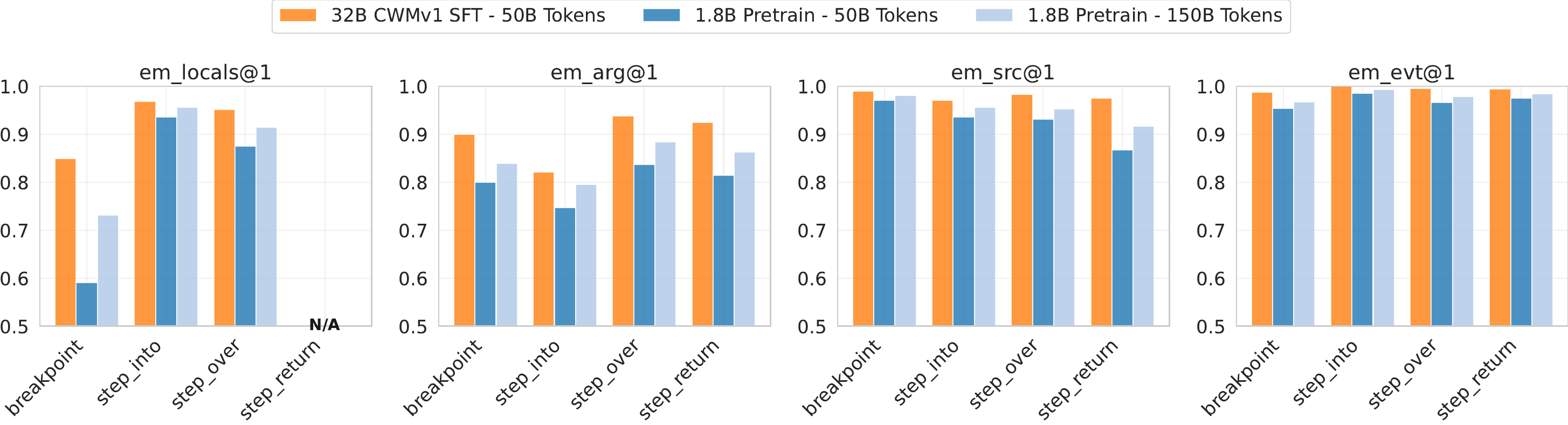}
        \label{fig:next_step_component_repolevel_forward}
    }
    \vfill
    \vspace{0.2cm}
    \subfloat[Inverse prediction.]{
        \includegraphics[width=0.8\linewidth]{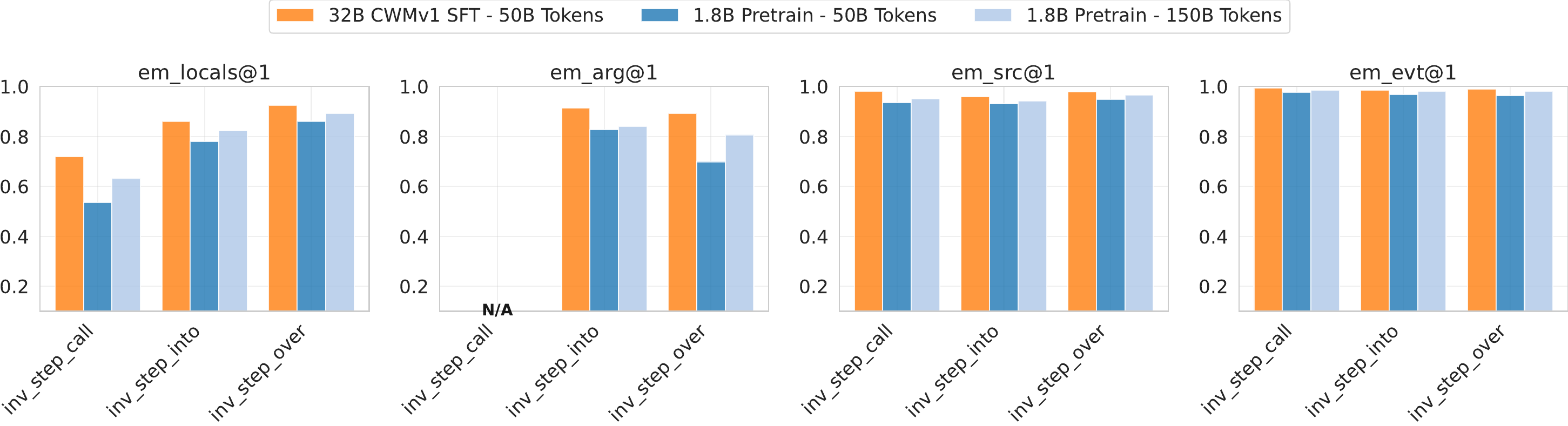}
        \label{fig:next_step_component_repolevel_inverse}
    }
    \caption{Exact match (em) next state prediction by state component per debugger action on the \textbf{repository-level} validation set. Figure~\ref{fig:next_step_component_funclevel} shows the corresponding results for the function-level validation set. Conclusions remain unchanged. \label{fig:next_step_component_repolevel}}
\end{figure}

In Figure~\ref{fig:next_step_component_repolevel}, we show the next state prediction results by state component per debugger action on the repository-level validation set.

\subsection{CruxEval input and output prediction prompts in neural debugger format}
\label{app:cruxeval_in_out_prompts}
In this section, we show examples of how we evaluate our neural debuggers on the CruxEval input and output prediction tasks.

In Figure~\ref{fig:cruxeval_output_prompts}, we show our prompts for CruxEval output prediction with the \texttt{step\_return} and \texttt{breakpoint} action.

\begin{figure}[h!]
  \centering
  \begin{subfigure}{0.47\textwidth}
    \lstset{style=simple_small}
    \begin{lstlisting}[language=Python]
<|begin_of_text|><|trace_context_start|>
def f(single_digit):
    result = []
    for c in range(1, 11):
        if c != single_digit:
            result.append(c)
    return result

def main():
    return f(5)
    
<|frame_sep|><|call_sep|>
<|src_sep|>def main():
<|arg_sep|>{}
<|action_sep|><|step_into|>

<|frame_sep|><|line_sep|>
<|arg_sep|>{}
<|src_sep|>    return f(5)
<|action_sep|><|step_into|>

<|frame_sep|><|call_sep|>
<|src_sep|>def f(single_digit):
<|arg_sep|>{"single_digit": "5"}
(*@\textcolor{cwm_string_color_bright}{<|action\_sep|><|step\_return|>}@*)

<|frame_sep|>
----END OF PROMPT----


<|return_sep|>
<|src_sep|>    return result
<|arg_sep|>"[1, 2, 3, 4, 6, 7, 8, 9, 10]"
    \end{lstlisting}
    \caption{Output prediction with \texttt{step\_return} action.}
  \end{subfigure}
  \hfill
  \begin{subfigure}{0.47\textwidth}
    \lstset{style=simple_small}
    \begin{lstlisting}[language=Python]
<|begin_of_text|><|trace_context_start|>
def f(single_digit):
    result = []
    for c in range(1, 11):
        if c != single_digit:
            result.append(c)
    return result

def main():
    return f(5)
    
<|frame_sep|><|call_sep|>
<|src_sep|>def main():
<|arg_sep|>{}
<|action_sep|><|step_into|>

<|frame_sep|><|line_sep|>
<|arg_sep|>{}
<|src_sep|>    return f(5)
<|action_sep|><|step_into|>

<|frame_sep|><|call_sep|>
<|src_sep|>def f(single_digit):
<|arg_sep|>{"single_digit": "5"}
(*@\textcolor{cwm_string_color_bright}{<|action\_sep|><|breakpoint|>    return result}@*)

<|frame_sep|>
----END OF PROMPT----

<|line_sep|>
<|arg_sep|>{"single_digit": "5", 
"result": "[1, 2, 3, 4, 6, 7, 8, 9, 10]", "c": "10"}
<|src_sep|>    return result
    \end{lstlisting}
    \caption{Output prediction with \texttt{breakpoint} action.}
  \end{subfigure}
  \caption{CruxEval output prompt for the \texttt{step\_return} and the \texttt{breakpoint} actions. 
  The outcome of the \texttt{step\_return} action is the frame with the return event and argument, while the outcome of the \texttt{breakpoint} action is the line event frame, including the local variables dictionary at the source line of the return statement.
  Predictions are generated by the \SI{1.8}{B} parameter neural debugger trained on \SI{150}{B} debugger trace tokens. Some line breaks are inserted for illustration purposes. \label{fig:cruxeval_output_prompts}}
\end{figure}

In Figure~\ref{fig:cruxeval_input_prompt}, we show the prompt for CruxEval input prediction with the \texttt{inv\_step\_call} action.

\begin{figure}[h!]
  \centering
  \begin{subfigure}{0.47\textwidth}
      \lstset{style=simple_small}
    \begin{lstlisting}[language=Python]
<|begin_of_text|><|trace_context_start|>
def f(single_digit):
    result = []
    for c in range(1, 11):
        if c != single_digit:
            result.append(c)
    return result
    
<|frame_sep|><|inv_return_sep|>
<|src_sep|>    return result
<|arg_sep|>"[1, 2, 3, 4, 6, 7, 8, 9, 10]"
(*@\textcolor{cwm_string_color_bright}{<|action\_sep|><|inv\_step\_call|>}@*)

<|frame_sep|>
----END OF PROMPT----

<|inv_call_sep|>
<|src_sep|>def f(single_digit):
<|arg_sep|>{"single_digit": "5"}
    \end{lstlisting}
\end{subfigure}
  \caption{CruxEval input prompt for the \texttt{inv\_step\_call} actions. Predictions are generated by the \SI{1.8}{B} parameter neural debugger trained on \SI{150}{B} debugger trace tokens. Some line breaks are inserted for illustration purposes. \label{fig:cruxeval_input_prompt}}
\end{figure}

\section{Neural debugger trace example}
\label{app:neural_debugger_examples}

In Figure~\ref{fig:debugger_trace_example_count_r}, we show the ground truth neural debugger trace with forward and inverse prediction. 
The example is inspired by the interactive Python code debugging example with CWM~\citep{codgenteam2025_cwm}.
In this case, the inverse debugger trace in Figure~\ref{fig:debugger_trace_example_count_r__inverse}is an example, where there are infinitely many possible \texttt{s} and \texttt{t} values for the return argument \texttt{n=2}.

\begin{figure}[h!]
  \centering
  \begin{subfigure}{0.52\textwidth}
    \lstset{style=simple_small_small}
    \begin{lstlisting}[language=Python]
<|begin_of_text|><|trace_context_start|>
def count(s, t):
    n = 0
    for c in s:
        n += int(c == t)
    return n

def main():
    return count("berry", "r")
<|frame_sep|><|call_sep|>
<|src_sep|>def main():<|arg_sep|>{}
<|action_sep|><|step_into|>

<|frame_sep|><|line_sep|>
<|arg_sep|>{}<|src_sep|>    return count("berry", "r")
<|action_sep|><|step_into|>

<|frame_sep|><|call_sep|>
<|src_sep|>def count(s, t):<|arg_sep|>{"s": "'berry'", "t": "'r'"}
<|action_sep|><|step_into|>

<|frame_sep|><|line_sep|>
<|arg_sep|>{"..": ".."}<|src_sep|>    n = 0
<|action_sep|><|step_into|>

<|frame_sep|><|line_sep|>
<|arg_sep|>{"..": "..", "n": "0"}<|src_sep|>    for c in s:
<|action_sep|><|step_into|>

<|frame_sep|><|line_sep|>
<|arg_sep|>{"..": "..", "c": "'b'"}<|src_sep|>        n += int(c == t)
<|action_sep|><|step_into|>

<|frame_sep|><|line_sep|>
<|arg_sep|>{"..": ".."}<|src_sep|>    for c in s:
<|action_sep|><|step_into|>

<|frame_sep|><|line_sep|>
<|arg_sep|>{"..": "..", "c": "'e'"}<|src_sep|>        n += int(c == t)
<|action_sep|><|step_into|>

<|frame_sep|><|line_sep|>
<|arg_sep|>{"..": ".."}<|src_sep|>    for c in s:
<|action_sep|><|step_into|>

<|frame_sep|><|line_sep|>
<|arg_sep|>{"..": "..", "c": "'r'"}<|src_sep|>        n += int(c == t)
<|action_sep|><|step_into|>

<|frame_sep|><|line_sep|>
<|arg_sep|>{"..": "..", "n": "1"}<|src_sep|>    for c in s:
<|action_sep|><|step_into|>

<|frame_sep|><|line_sep|>
<|arg_sep|>{"..": ".."}<|src_sep|>        n += int(c == t)
<|action_sep|><|step_into|>

<|frame_sep|><|line_sep|>
<|arg_sep|>{"..": "..", "n": "2"}<|src_sep|>    for c in s:
<|action_sep|><|step_into|>

<|frame_sep|><|line_sep|>
<|arg_sep|>{"..": "..", "c": "'y'"}<|src_sep|>        n += int(c == t)
<|action_sep|><|step_into|>

<|frame_sep|><|line_sep|>
<|arg_sep|>{"..": ".."}<|src_sep|>    for c in s:
<|action_sep|><|step_into|>

<|frame_sep|><|line_sep|>
<|arg_sep|>{"..": ".."}<|src_sep|>    return n
<|action_sep|><|step_into|>

<|frame_sep|><|return_sep|><|src_sep|>    return n
<|arg_sep|>"2"
<|action_sep|><|step_into|>

<|frame_sep|><|return_sep|>
<|src_sep|>    return count("berry", "r")<|arg_sep|>"2"
<|action_sep|><|step_into|>

<|frame_sep|><|exit_normal|><|trace_end|><|end_of_text|>
    \end{lstlisting}
    \caption{Forward debugger trace.}
  \end{subfigure}
  \hfill
  \begin{subfigure}{0.47\textwidth}
    \lstset{style=simple_small_small}
    \begin{lstlisting}[language=Python]
<|begin_of_text|><|trace_context_start|>
def count(s, t):
    n = 0
    for c in s:
        n += int(c == t)
    return n
<|frame_sep|><|inv_return_sep|>
<|src_sep|>    return n<|arg_sep|>"2"
<|action_sep|><|inv_step_into|>

<|frame_sep|><|inv_line_sep|>
<|src_sep|>    return n<|arg_sep|>{"..": ".."}
<|action_sep|><|inv_step_into|>

<|frame_sep|><|inv_line_sep|>
<|src_sep|>    for c in s:<|arg_sep|>{"..": ".."}
<|action_sep|><|inv_step_into|>

<|frame_sep|><|inv_line_sep|>
<|src_sep|>        n += int(c == t)<|arg_sep|>{"..": ".."}
<|action_sep|><|inv_step_into|>

<|frame_sep|><|inv_line_sep|>
<|src_sep|>    for c in s:<|arg_sep|>{"..": "..", "c": "'r'"}
<|action_sep|><|inv_step_into|>

<|frame_sep|><|inv_line_sep|>
<|src_sep|>        n += int(c == t)<|arg_sep|>{"..": "..", "n": "1"}
<|action_sep|><|inv_step_into|>

<|frame_sep|><|inv_line_sep|>
<|src_sep|>    for c in s:<|arg_sep|>{"..": ".."}
<|action_sep|><|inv_step_into|>

<|frame_sep|><|inv_line_sep|>
<|src_sep|>        n += int(c == t)<|arg_sep|>{"..": "..", "n": "0"}
<|action_sep|><|inv_step_into|>

<|frame_sep|><|inv_line_sep|>
<|src_sep|>    for c in s:<|arg_sep|>{"..": "..", "c": "'e'"}
<|action_sep|><|inv_step_into|>

<|frame_sep|><|inv_line_sep|>
<|src_sep|>        n += int(c == t)<|arg_sep|>{"..": ".."}
<|action_sep|><|inv_step_into|>

<|frame_sep|><|inv_line_sep|>
<|src_sep|>    for c in s:<|arg_sep|>{"..": "..", "c": "'b'"}
<|action_sep|><|inv_step_into|>

<|frame_sep|><|inv_line_sep|>
<|src_sep|>        n += int(c == t)<|arg_sep|>{"..": ".."}
<|action_sep|><|inv_step_into|>

<|frame_sep|><|inv_line_sep|>
<|src_sep|>    for c in s:<|arg_sep|>{"..": ".."}
<|action_sep|><|inv_step_into|>

<|frame_sep|><|inv_line_sep|>
<|src_sep|>    n = 0<|arg_sep|>{"..": ".."}
<|action_sep|><|inv_step_into|>

<|frame_sep|><|inv_call_sep|>
<|src_sep|>def count(s, t):<|arg_sep|>{"s": "'berry'", "t": "'r'"}
<|action_sep|><|inv_step_into|>

<|frame_sep|><|inv_exit_entry|><|trace_end|><|end_of_text|>
    \end{lstlisting}
    \caption{Inverse debugger trace. \label{fig:debugger_trace_example_count_r__inverse}}
  \end{subfigure}
  \caption{An example of a forward and inverse debugger trace for a function counting the occurrences of \texttt{t} in \texttt{s}. We take only \texttt{step\_into} or \texttt{inv\_step\_into} actions to visit every frame. While the forward debugger trace is deterministic given its inputs, the inverse debugger trace with return argument \texttt{n=2} is an example with infinitely many input combinations for \texttt{s} and \texttt{t}. \label{fig:debugger_trace_example_count_r}}
\end{figure}

\end{document}